\documentclass[twoside,11pt]{article}
\usepackage{jair, theapa, rawfonts}
\usepackage[ruled,vlined]{algorithm2e}
\usepackage[export]{adjustbox}

\ShortHeadings{Learning  Realistic Patterns from Visually Unrealistic Stimuli}
{Nikolaidis et al.}
\firstpageno{1}

\usepackage{subcaption}

\usepackage{wrapfig}
\usepackage{graphicx}
\usepackage{amsfonts}
\usepackage{amsmath}

\newtheorem{theorem}{Lemma}

\begin{document}

\title{Learning  Realistic Patterns from Visually Unrealistic Stimuli: Generalization and Data Anonymization}

\author{\name  Konstantinos Nikolaidis \email konstan@ifi.uio.no \\
        \name  Stein Kristiansen \email steikr@ifi.uio.no \\
        \name  Thomas Plagemann \email plageman@ifi.uio.no\\
        \name  Vera Goebel \email goebel@ifi.uio.no \\
        \name  Knut Liestøl \email knut@ifi.uio.no \\
        \addr Department of Informatics\\ University of Oslo, Norway\\\\
          \name   Mohan Kankanhalli \email mohan@comp.nus.edu.sg \\
        \addr Department of Computer Science \\
        National University of Singapore, Singapore\\\\
          \name   Gunn Marit Traaen \email gunnmarit.traaen@gmail.com \\
        \addr Department of Cardiology\\ Oslo University Hospital Rikshospitalet, Oslo, Norway\\\\
        \name   Britt Øverland \email britt.overland@lds.no \\
        \addr Department of Otorhinolaryngology, Head \& Neck Surgery\\ Sleep Unit Lovisenberg Diakonale Hospital, Oslo, Norway\\\\
          \name   Harriet Akre \email harriet.akre@gmail.com\\
        \addr Institute of Clinical Medicine\\ Faculty of Medicine, University of Oslo, Oslo, Norway\\
        \addr Department of Otorhinolaryngology, Head \& Neck Surgery\\ Oslo University Hospital, Rikshospitalet, Oslo, Norway\\\\
          \name  Lars Aakerøy \email lars.aakeroy@stolav.no \\
          \name  Sigurd Steinshamn\email sigurd.steinshamn@ntnu.no \\
        \addr Department of Circulation and Medical Imaging\\ Faculty of Medicine and Health Science\\ Norwegian University of Science and Technology, Trondheim, Norway\\\\}


\maketitle

\begin{abstract}
Good training data is a prerequisite to develop useful Machine Learning applications. However, in many domains existing data sets  cannot be shared due to privacy regulations (e.g., from medical studies). This work investigates a simple yet unconventional approach for anonymized data synthesis to enable third parties to benefit from such anonymized data. We explore the feasibility of learning implicitly from visually unrealistic, task-relevant stimuli, which are synthesized by exciting the neurons of a  trained  deep neural network. As such, neuronal excitation can be used to generate synthetic stimuli.  The stimuli data is used to train new classification models. Furthermore, we extend this framework to inhibit representations that are associated with specific individuals. We use sleep monitoring data from both an open and a large closed clinical study, and Electroencephalogram sleep stage classification data, to evaluate whether (1) end-users can create and successfully use customized classification models, and (2) the identity of participants in the study is protected. Extensive comparative empirical investigation  shows that different algorithms trained on the stimuli are able to generalize successfully on  the same task as the original model.   Architectural and algorithmic similarity between  new and  original models  play an important role in performance.  For similar architectures, the performance is close to that of  using the original data (e.g., Accuracy difference of 0.56\%-3.82\%, Kappa coefficient difference of 0.02-0.08).  Further experiments show that the stimuli can     provide state-of-the-art resilience against adversarial association and membership inference attacks. 
\end{abstract}

\section{Introduction}

In recent years, machine learning (ML) has become a viable solution for various applications due to rapid developments in sensor technologies, data acquisition tools, and ML algorithms (e.g., deep learning). It is well-known that training data of sufficient quality and quantity is a pre-requisite to train a ML classification model (classifier), that can generalize reliably. However, there are situations in which access to data that fulfil these requirements is restricted, e.g., due to privacy concerns.

Such situations are particularly prominent in the medical domain. One example is the Cesar project \shortcite{kristiansen2021clinical}, which aims to enable individuals to perform sleep monitoring at home with low-cost sensors and ML-based automatic sleep apnea detection on their smart-phone. Having access to labelled sensor data from a large clinical study enables us in the project to train ML classification models and evaluate their performance. The final goal of the project is to allow any individual to use a customized classifier that is tailored to the particular needs of the individual,  e.g., to use it in a resource constrained environment. However, regulatory restrictions prohibit us to share the data, neither with individuals so that they can create their own customized classifiers, nor for other scientific purposes.  Creating a customized classifier for any interested individual in our lab is not a
feasible solution.

A possible solution to this problem is to release an anonymized version of the data with the use of existing database anonymization strategies like k-anonymity \shortcite{sweeney2002k} or l-diversity \shortcite{machanavajjhala2007diversity}. However, since crucial parts of our data are raw sensory time-series data from which identification can be done indirectly via learning, existing database anonymization strategies are not suitable for this task. Another option would be to only release white-box or black-box classifiers in the form of an API or mechanisms that perform classification or extract important statistics from the data. For this purpose, differential privacy  (DP) is a well-established framework that offers theoretical guarantees for privacy preserving application of statistical mechanisms \shortcite{dwork2008differential}. Works such as by Abadi et al. \citeyear{abadi2016deep} make DP a viable option for ML applications. However, this option defeats our goal of giving customization freedom to the end-users. Additionally, though theoretically sound, DP has been shown to yield in many cases unacceptable privacy-performance trade-offs \shortcite{jayaraman2019evaluating,fredrikson2014privacy}, and to be susceptible to different information leakage attacks \shortcite{zhang2019secret,hitaj2017deep}. 
A third option is to train a differentially private generative model, like the ones proposed   by  Jordon et al. \citeyear{jordon2018pate} or Xie et al. \citeyear{xie2018differentially},   on our data,  in order to synthesize a dataset that we could release to the public. We avoid this option for two reasons: (1) because of the points discussed above about DP; (2) as we care about the identification of recordings and not of individual datapoints, we would need to apply the group privacy property while training such a model. However, this would weaken either the attainable privacy guarantee or the performance.

Based on the above discussion, we explore in this work a different option, which is not sufficiently studied in related works. We investigate the empirical feasibility of labelled noisy higher-layer representations for training other ``student'' classifiers to generalize reliably on the real data. The goal is to create a labelled dataset from which a model can be taught to perform   classification, while at the same time data from this dataset cannot be strongly identified as belonging to a specific recording. To do this, we exploit the knowledge obtained by a given trained classifier, which we refer to as Teacher, $h_T$ for notational convenience. $h_T$ is trained to capture the most important aspects of the real data based on the loss it attempts to minimize, making it learn task-related knowledge about the training data. We expect that excitatory or inhibitory datapoints (which we call \textit{stimuli}) resulting from the activation of specific neurons can also contain important information about the class decisions of $h_T$. Based on this, we learn to generate varying stimuli targeting the output of one or more neurons of $h_T$ and use these stimuli to train a student classifier $h_S$. \textit{Neuronal Excitation} (NE), is a general method that can be applied on artificial neural networks \shortcite{Nguyen_2015_CVPR} as well as on the mammalian inferotemporal cortex \shortcite{ponce2019evolving}.   In this work we use   \textit{Activation Maximization} or Minimization (AM) \shortcite{erhan2009visualizing} as NE.

\subsection{Activation Maximization}

AM is a simple  method which searches for input patterns that maximize the activation of a hidden or output unit  of a Deep Neural Network (DNN). The reasoning behind this idea is that \textit{``a pattern to which the unit is responding
maximally could be a good first-order representation of what a unit is doing''} \cite{erhan2009visualizing}. A simple way to  find such data is to identify training or test datapoints which activate maximally the hidden unit we are interested in. However, a more general view can be established and  the problem can be reduced into an optimization problem of the input space. As such, gradient ascent can be performed into the input space, such that the output of the hidden neuron we are interested in (in our case one of the output neurons of $h_T$) can be maximized. Through this procedure we can identify an appropriate datapoint which adheres to our requirement, and maximizes the neuronal unit we are interested in. If the gradient ascent converges appropriately this datapoint will probably be a local maximum.

 We extend this idea as follows: if $h_T$ is trained, then multi-faceted stimuli that occur from  NE into the ouput neurons of $h_T$ can contain sufficient knowledge about the class separation of the task we are interested in.

\subsection{Unrealistic Data}
 The overall proposed procedure of this work is loosely related to implicit learning \shortcite{reber1989implicit} in the sense that for $h_S$, knowledge about features of the true joint distribution is acquired implicitly, and not through direct loss minimization on data sampled from it (or from a distribution that approximates it).
This leads us to an important novel aspect of this work that to the best of our knowledge serves as a differentiating factor compared to other generative approaches. The stimuli we are synthesizing need not necessarily be realistic. On the contrary, to some extent we want them to be unrealistic. We want $h_S$ to learn indirectly through the stimuli and generalize on the real data. Please note that such an approach only captures those features needed to strongly excite or inhibit different class neurons of $h_T$. This is in contrast to a generative model or framework which would attempt to capture all features necessary to learn the joint or  the marginal distribution based on its loss. Therefore, we have more direct access to the conditional distribution  we want to learn. We hypothesize that this procedure is a natural way to generate datapoints that, though visually unrealistic,   contain inherently important information about the class separation we care about, e.g., sleep apnea in our case. Additionally, the datapoints   potentially provide less ``unwanted'' information for other class separations which we want not to be learned. 

\subsection{Contributions}
In this work, we  empirically investigate the viability of the proposed approach through several case studies with real-world health data. Our contributions are as follows:
 \begin{itemize}
     \item We demonstrate that learning from AM generated stimuli is an empirically feasible way to learn and generalize successfully on new data.
     \item We investigate the applicability of training different smaller architectures for successful customization with the use of the generated stimuli dataset. We compare with two existing well-established generative approaches, namely gradient-penalty Wasserstein Generative Adversarial Network (GAN) \shortcite{pmlr-v70-arjovsky17a} and Variational Autoencoders (VAE) \shortcite{kingma2013auto}, and provide promising results. The performance obtained when using AM stimuli is close to the performance obtained when using the original data, with differences ranging from 0.025 to 0.082  in terms of Kappa  coefficient, and  from 0.56\% to 3.82\% in terms of Accuracy, depending on dataset and configuration.
     \item We empirically show the viability of a variation of the proposed approach as a means of generating anonymized data. To do this, we develop a patient de-anonymization attack inspired from face identification. We evaluate how the AM stimuli compare to the real data in terms of the identification success of the adversary  and investigated task performance. Furthermore, we evaluate how differentially private variants of the generative models, and the CFUR \shortcite{kairouz2019censored} generative de-anonymization strategy perform and showcase state-of-the-art comparative results (i.e., Kappa improvements in de-identification  ranging from 0.02 to 0.41).
     
     \item Finally, we explore the defence capability that the proposed approach offers against membership inference attacks and exhibit additional potentially useful properties of the described method.
 \end{itemize}
 
 \subsection{Paper Structure}
The rest of the paper is organized as follows: Section \ref{Method} presents the proposed approach. Section \ref{sec:Datasets} describes the application scenario and the datasets we use. Section \ref{sec:prelim} discusses preliminary insights of the proposed approach. In Section \ref{sec:genandcust}, 
we perform experiments to investigate the generalization and customization capabilities of the proposed approach. In Section \ref{sec:defandanon}, we investigate defensive and anonymization properties and   in Section \ref{sec:Discussion1} we  discuss additional characteristics and limitations of the proposed  approach. Section \ref{sec:rw} compares   with related literature. Section \ref{sec:concl} concludes this paper \footnote{We include relevant proof-of-concept code at: https://github.com/konkoniknik/AM-Stimuli/}.

\section{Approach}
\label{Method}
In this section, we discuss the approach we use to generate a synthetic labelled dataset consisting of  stimuli via the use of AM. We describe first the overall idea, followed by an analysis of the core components of the approach. Furthermore, we discuss  extensions that allow for anonymized data generation and   how  AM generated stimuli can provide an efficient way to learn.  Finally, we give a working example of the whole procedure to make it more understandable.

\begin{figure}[h]
\centering
  \includegraphics[scale=0.47]{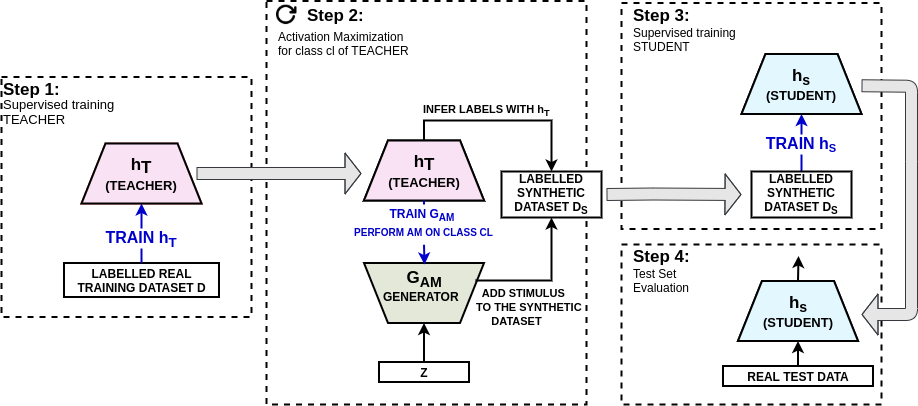}
\caption{ Four main steps of the proposed approach. TEACHER corresponds to $h_T$ and STUDENT to $h_S$. The GENERATOR ($G_{AM}$) performs AM on a randomly chosen class of $h_T$ (class $cl$). The procedure is repeated multiple times with randomly chosen classes of $h_T$ to generate the stimuli dataset.}
\label{fig:Method}
\end{figure}

\subsection{Overview}
\label{Method2}

We want to transfer the knowledge of a given trained DNN classifier  $h_T$ to another model or learning algorithm  $h_S$ through the use of a synthetic dataset $D_S$, which  sufficiently captures this knowledge. We define the input space $\mathbb X$ as a compact subset of $\mathbb R^K$ where $K$ is the input dimensionality. Furthermore, we define the output space $\mathbb Y=\{0,...,N_c-1\}$,  which corresponds to the different classes of the classification task, where $N_c$ is the number of classes. Both  classifiers $h_S$ and $h_T$ have the same input space $\mathbb X$ and output that is normalized into a probability distribution, e.g., with the use of a softmax activation, with as many values as the number of different classes in $\mathbb Y$. As such, we assume that $\sum_{c=0}^{N_c-1} h^{(c)}_T=\sum_{c=0}^{N_c-1} h^{(c)}_S=1$ and $\arg\max_c h^{c}_T,\arg\max_c h^{c}_S : \mathbb X\rightarrow \mathbb Y$. Furthermore,  we assume that the original labelled data $D$,  with which $h_T$ is trained is not available for training of $h_S$ \footnote{In this work we use the terms \textit{real} data and \textit{original} data interchangeably to describe $D$.}. The end-user who trains $h_S$ only has access to $D_S$.  We aim to enable $h_S$ to classify data that come from the same distribution as $D$ with a similar performance as $h_T$. One way to do this efficiently is to extract the knowledge accumulated by $h_T$ and create a synthetic dataset $D_S$ that represents this knowledge. The novelty of the proposed approach stems from the fact that we utilize AM in an unconventional manner with the goal of creating a diverse, multi-faceted dataset  $D_S$ that can be used to train another student classifier.

The achievable classification performance of $h_S$ depends on the one hand on the success of the generation procedure to map the important features learned by $h_T$ onto $D_S$, and on the other hand on the algorithmic and architectural similarity between $h_T$ and $h_S$.

\par \paragraph{Design:} Our proposed design is based on four basic steps (see Figure \ref{fig:Method}):
\begin{itemize}
\item \textbf{Step 1} - Training of the teacher. We train $h_T$ in a supervised manner with  $D$  to learn the underlying conditional  distribution of task $\mathbb X \rightarrow\mathbb Y$. This requires the original labelled training data.
\item \textbf{Step 2} - Creating $D_S$. We create a synthetic dataset $D_S$ that captures features that $h_T$ has learned from training on $D$. We perform AM via a deep generator network $G_{AM}$. Inspirations for this design are \shortcite{baluja2017adversarial,nguyen2016synthesizing}. $G_{AM}$ is optimized to transform   input random noise vectors into  stimuli that strongly activate a  pre-chosen output class neuron $cl$ of $h_T$. After $G_{AM}$'s optimization is finished, we use $G_{AM}$ to generate one or multiple synthetic stimuli. This process is repeated multiple times to create multiple stimuli for $D_S$. Each time an output class neuron of $h_T$ is randomly selected. After the synthetic stimuli are created, we create their labels. To do this, we pair each   chosen synthetic stimuli  with its corresponding output from $h_T$. 
\item \textbf{Step 3} - Training of the student. As next step, $h_S$ can be trained with the synthetic data and labels produced by Step 2. $h_S$ can be a larger or smaller DNN than $h_T$, or even be based on a different learning method, e.g, an SVM.
\item \textbf{Step 4} - Evaluate $h_S$ with the test set. 
\end{itemize}

Next, we discuss the basic components of the proposed approach in more detail.  Steps 1, 3, and 4 are either trivial or represent already known methodologies which are commonly used in ML literature. As such, we discuss briefly these steps in Appendix H, and we focus our main analysis  on Step 2.

\subsection{Step 2: Generating $D_S$}

For Step 2, we use the trained classifier $h_T(\theta^*_T,\cdot)$ obtained from Step 1, where $\theta^*_T \in \Theta_T$ correspond to the optimized parameters of $h_T$, and $\Theta_T$ is the parameter space of $h_T$. We create $D_S$ through repeated AM on different randomly selected output neurons of $h_T$.
The only necessary  component we need for Step 2 is a trained gradient-based classifier. As such this Step can take place separately from Step 1, given that we have the classifier. 

\subsubsection{ Optimization and Choosing the Stimuli} 
We use a generator network $G_{AM}$ to transform a random noise vector $z$  drawn from  a  prior distribution with density $p(z)$, into a stimulus that strongly activates a neuron of $h_T$. We focus  on output neurons of $h_T$ which correspond to different classes. 

We perform the AM as follows: we randomly select  (equal probability) one of the output class neurons. We assume that the selected neuron corresponds to class $cl$.  Then, for a mini-batch of prior samples, we minimize the categorical cross-entropy between the one-hot encoding for the chosen class  $\mathbf{y_{cl}}=onehot\{cl\}$  and $h_T(\theta^*_T,G_{AM}(\theta_G,z))$\footnote{We assume row vectors}. Thus, for a mini-batch of size $m$ of $\{z_i\}_{i=1}^m \sim p(z)$ the loss $L_Y(\theta_G)$ can be expressed as follows:

\begin{equation}
\begin{split}
    L_Y(\theta_G) & = -\frac{1}{m}\sum_{i=0}^m \sum^{N_c-1}_{c=0} \mathbf{y_{cl}}^{(c)}  \log h^{(c)}_T(\theta^*_T,G_{AM}(\theta_G,z_i))\\
     & = -\frac{1}{m}\sum_{i=0}^m   \log h^{(cl)}_T(\theta^*_T,G_{AM}(\theta_G,z_i))\\
\end{split}
\end{equation}

 assuming $h_T$ has $N_c$ output class neurons, and the superscript $(c)$ indexes the element corresponding to class $c$. The objective is to minimize $L_Y(\theta_G)$:

\begin{equation}
\theta^*_G=\arg\min_{\theta_G} \{L_Y(\theta_G)\}
\end{equation}

This process is repeated many times, with random reselection of $cl$ to form $D_S$ with the outputs of $G_{AM}$. The optimization  only affects $\theta_G$, and $\theta^*_T$ always remains unchanged. To train $G_{AM}$, we do not use only one mini-batch of $p(z)$ samples, but instead   a new minibatch is resampled per optimization step. Furthermore, $p(z)$ is resampled again to generate candidate samples that are   potentially to be included in $D_S$. We perform these steps  to better fit the transformation of $z$ via $G_{AM}$ and to  provide better regularization.

To choose which of the candidate stimuli samples are to be included in $D_S$, we examine which of these strongly activate the pre-chosen output neuron. If a stimulus from this candidate set provides an activation that surpasses a threshold $T_Y$ we include it in $D_S$. Formally, for a stimuli $x_s=G_{AM}(\theta_G,z)$ we want: 

\begin{equation}
    h^{(cl)}_T(\theta^*_T,x_s)>T_Y
    \label{eq:Thresh}
\end{equation}

and if this condition holds, then the pair $(x_s,h_T(\theta^*_T,x_s))$ is included in $D_S$. We perform optimization steps for the selected neuron until $m'$  stimuli pass the condition or until a maximum number of optimization steps $n_{mx}$ is reached. The whole procedure is shown in Algorithm 1. 

Based on this discussion, the threshold value $T_Y$, and to an extent, the size of the optimization and candidate mini-batches play an important role in the overall procedure. We discuss design and value decisions for these hyperparameters in Section 2.6.

\subsubsection{Prior Distribution $p(z)$}

The prior samples $z$ which are being transformed into stimuli from $G_{AM}$ have in the majority of our experiments lower dimensionality than the stimuli. Furthermore, we use as $p(z)$ from which we draw prior samples $z$ the random normal distribution with mean 0 and a  standard deviation of 1. The dimensionality of $z$   changes depending on the experiment.

\begin{algorithm}
\small
\SetAlgoLined
\KwIn{Teacher Model: $h_T$, Number of classes: $N_c=dim(\mathbb Y)$, AM-Generator: $G_{AM}$, Prior Distribution: $p(z)$, Batch Size: $m$, Target size of stimuli dataset: $N_S$, Acceptance Threshold: $T_Y$,  Maximum number of Added Stimuli per AM: $m'$, learning rate: $l$    }

$D_S\leftarrow\emptyset$\;
Specify a break limit for AM: $n_{MX}$\;
\While{ $|D_S|<N_S$}{
Select output class neuron randomly: $cl\leftarrow Random(N_c)$\;

$D_t\leftarrow \emptyset$,
$n\leftarrow0$\;
Reinitialize $G_{AM}$: $\theta_G^{(n)}\leftarrow Init\{G_{AM}\}$\;

\While{$|D_t|\leq m'$ and $n<n_{MX}$  }{

    Sample a mini-batch of prior samples: $\{z_i\}_{i=1}^m \sim p(z)$\;
    $L_Y(\theta^{(n)}_G)\leftarrow -\frac{1}{m}\sum_{i=1}^m \mathbf{y_{cl}} \cdot (\log h_T^{(c)}(\theta^*_T,G_{AM}(\theta^{(n)}_G,z_i)))_c^\top $\;
 Update  $G_{AM}$:  $\theta_G^{(n+1)}\leftarrow  Optimize\{L_Y,\theta_G^{(n)},l\}$\;
    $n\leftarrow n+1$\;

    Sample a new mini-batch of prior samples: $\{z_j\}_{j=1}^{m'} \sim p(z)$\;
    \For {all $z_j$}{
    Create stimulus for $z_j$: $x_{s_j}\leftarrow G_{AM}(\theta_G^{(n)},z_j))$\;
    Create stimulus labels: $\mathbf{sl}_j\leftarrow h_T(\theta^*_T,x_{s_j})$\;
    \If{$\mathbf{sl}_j^{(cl)}>T_Y$}{$D_t\leftarrow D_t\cup (x^{(j)}_s,\mathbf{sl}_j)$}
    }
}
$D_S\leftarrow D_S\cup D_t$
}
 \KwOut{Stimuli Dataset: $D_S$}
 \caption{AM-based Stimuli Generation}
\end{algorithm} 

\subsubsection{Adding Randomness and Diversity}

To make the generated synthetic stimuli more diverse, i.e., to increase the variety of input space positions of the stimuli, we add inherent randomness in the AM procedure. We do this for example, by reinitializing the AM generator, or by randomly selecting $T_Y$ from several possible values, among other techniques (for more details,  please refer to Appendix B).  Our goal is to capture a wide variety of different starting positions in $\mathbb X$ for the gradient descent. Additionally, we want to randomize where the optimization will end up, given that the threshold condition is satisfied. We perform the AM from each initial position towards a random class with equal probability to create a stimulus. The result is a synthetic dataset comprised of the stimuli of $h_T$ for all the classes. The goal of the different initial input space positions is to take advantage of the multi-faceted property of the neuronal activation \shortcite{nguyen2016multifaceted} when performing AM to create a diverse synthetic dataset, that can offer implicitly sufficient information for the conditional data distribution.

\begin{figure}[h]
\centering
  \includegraphics[scale=0.5]{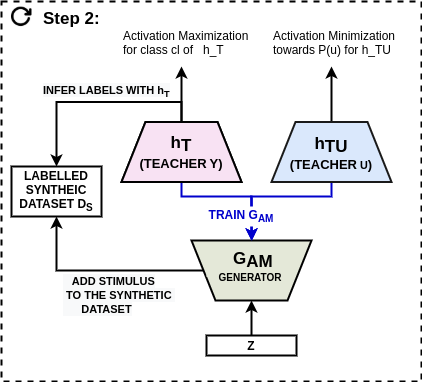}
\caption{ Extension of Step 2: Synthesizing stimuli that excitate the conditional  $(p(y|x))_{y\in\mathbb Y}$ approximator $h_T(x)$, depicted as TEACHER$_Y$,while inhibiting the $(p(u|x))_{u\in\mathbb U}$ approximator $h_{TU}(x)$, depicted as TEACHER$_U$. This extension replaces Step 2 of Figure 1.  }
\label{fig:MethodExcInh}
\end{figure}

\subsubsection{Recording Anonymization with Inhibitory Stimuli }
\label{sec:Extension}

To satisfy our objective of generating anonymized stimuli, we need to extend our approach. We want to   generate stimuli which  do not give away characteristic features that belong to specific individuals in the original data. Our proposed extension is loosely inspired  by the analysis of Feutry et al. \citeyear{feutry2018learning}. We assume a new learning task $\mathbb X\rightarrow \mathbb U$, different  from the original $\mathbb X\rightarrow \mathbb  Y$. Assuming that $N_p$ is the number of individuals that contributed to the original dataset $D$, we create $\mathbb U$ by assigning a unique number from  $\{0,..,N_p-1\}\in\mathbb{N}$  to each individual that contributed to $D$.
For example, assuming an image dataset of faces comprised of 3   individuals, e.g., John, Mary,  and Tom each contributing a certain number of images, $N_p=3$ and $\mathbb U=\{0,1,2\}$, with each natural number corresponding to one of these individuals (i.e., $0\rightarrow$``John'', $1\rightarrow$``Mary'', $2\rightarrow$``Tom'').  Each datapoint in $D$ is assigned the  label of the individual which it originates from.

 To model this labelling, we use a discrete random variable (r.v) $U\in \mathbb U$, conditionally dependent on r.v $X\in \mathbb X$ by $P(U=u|X=x)=p(u|x)$. Our goal is to generate data that are not strongly affiliated with any value of $U$. Formally, we want: $\forall x_s\in D_S, \forall u\in\mathbb U: p(u|x_s)$ to be close to $p(u)$, where $p(u)$ is the marginal distribution of $U$, i.e., $P(U=u)=p(u)$. We hypothesize that if r.v $Y\in \mathbb Y$ and $U$ are not strongly correlated given different values of $X$ then learning for $\mathbb X\rightarrow \mathbb Y$ would be possible with data that satisfy the previous requirement.

\begin{algorithm}[H]
\small
\SetAlgoLined
\KwIn{Teacher $Y,U$ Model: $h_T$, $h_{TU}$, Number of classes: $N_c=dim(\mathbb Y)$, AM-Generator: $G_{AM}$, Prior Distribution: $p(z)$, Batch Size: $m$, Size of $D_S$: $N_S$, Acceptance Threshold $Y$,$U$: $T_Y,T_U$,  Maximum  Number of Added Stimuli per AM: $m'$, learning rates $Y,U$ : $l_Y,l_U$, Dataset $D$    }

$D_S\leftarrow\emptyset$\;
Specify a break limit for AM: $n_{MX}$\;
Calculate estimate for marginal $p(u)\forall u\in \mathbb U$ based on  $D$: $(\hat{p}(u))_{u\in \mathbb U}$

\While{ $|D_S|<N_S$}{
Select output class neuron randomly: $cl\leftarrow Random(N_c)$\;

$D_t\leftarrow \emptyset$,
$n\leftarrow0$\;
Reinitialize $G_{AM}$: $\theta_G^{(n)}\leftarrow Init\{G_{AM}\}$\;

\While{$|D_t|\leq m'$ and $n<n_{MX}$  }{

    Sample a mini-batch of prior samples: $\{z_i\}_{i=1}^m \sim p(z)$\;
    $L_Y(\theta^{(n)}_G)\leftarrow -\frac{1}{m}\sum_{i=1}^m \mathbf{y_{cl}} \cdot (\log h_T^{(c)}(\theta^*_T,G_{AM}(\theta^{(n)}_G,z_i)))_c^\top $\;
     Update  $G_{AM}$, Task $Y$:  $\theta_G^{(n+1)}\leftarrow  Optimize\{L_Y,\theta_G^{(n)},l_Y\}$\;
    $L_U(\theta^{(n+1)}_G)\leftarrow -\frac{1}{m}\sum_{i=1}^m (\hat{p}(u))_u \cdot (\log h_{TU}^{(c)}(\theta^*_{TU},G_{AM}(\theta^{(n+1)}_G,z_i)))_c^\top $\;

 Update  $G_{AM}$, Task $U$:  $\theta_G^{(n+2)}\leftarrow  Optimize\{L_U,\theta_G^{(n+1)},l_U\}$\;
 
    $n\leftarrow n+2$\;

    Sample a new mini-batch of prior samples: $\{z_j\}_{j=1}^{m'} \sim p(z)$\;
    \For {all $z_j$}{
    Create stimulus for $z_j$: $x_{s_j}\leftarrow G_{AM}(\theta_G^{(n)},z_j))$\;
    Create stimulus labels: $\mathbf{sl}_j\leftarrow h_T(\theta^*_T,x_{s_j})$\;
    \If{($\mathbf{sl}_j^{(cl)}>T_Y$)and($L_U(\theta^{(n)}_G)< T_U$)}{$D_t\leftarrow D_t\cup (x_{s_j},\mathbf{sl}_j)$}
    }
}
$D_S\leftarrow D_S\cup D_t$
}
 \KwOut{Stimuli Dataset: $D_S$}
 \caption{AM-based Stimuli Generation with Recording  Inhibition}
\end{algorithm}

In practice, we achieve this as follows: we train a network $h_{TU}(x)$ to approximate the conditional distribution vector $(p(u|x))_{u\in\mathbb U}$ in $\mathbb X$, and learn to identify for each data point the individual it belongs to. Then, for all $x_s \in D_S$ we minimize the cross-entropy between $h_{TU}(x_s)$  and $(\hat p(u))_{u\in\mathbb U}$. $\hat{p}(u)$  is the empirical approximation of the marginal distribution of $u$.  This means that we minimize the cross entropy between    the $p(u|x)$ approximator,  and  the approximation of the marginal distribution $p(u)$, which we calculate based on $D$.  We combine this objective with the original AM objective and alternate training updates between them. This means that for a mini-batch of size $m$ of $\{z_i\}_{i=1}^m \sim p(z)$, we alternate  between   training  updates for the original objective   and  the following cross-entropy loss:
\begin{equation}
    L_U(\theta_G)  = -\frac{1}{m}\sum_{i=1}^m \sum^{N_p-1}_{u_c=0} \hat{p}(u_c)\log {h^{(u_c)}_{TU}(\theta^*_{TU},G_{AM}(\theta_G,z_i))}  
\end{equation}
   where $\theta^*_{TU}\in \Theta_{TU}$ are the parameters of $h_{TU}$, and $\Theta_{TU}$ its parameter space. The optimization is  done  for $\theta_G$, after training $h_T$ and $h_{TU}$. The parameters of $h_T$ and $h_{TU}$ are frozen. To calculate each $\hat{p}(u)$, and form vector $(\hat p(u))_{u\in \mathbb U}$, we calculate the empirical probability mass per individual. We do this by counting the datapoints that belong to each individual and dividing by the total amount of datapoints in $D$. In the above example, if the dataset contains 150 datapoints and John contributed 100, Mary 20 and Tom 30 images, then we would have: $\hat p(u=0)=\frac{100}{150},\hat p(u=1)=\frac{20}{150}, \hat p(u=2)=\frac{30}{150}$, and thus $(\hat p(u))_{u\in\mathbb U}=[\frac{100}{150},\frac{20}{150},\frac{30}{150}]$. The optimization stops when  the threshold condition of Eq. \ref{eq:Thresh} and a proximity condition regarding $(\hat{p}(u))_{u\in \mathbb U}$ are satisfied (e.g., $L_U(\theta_G)<T_U$, for some threshold $T_U$, which we discuss in Section 2.6). The whole procedure is outlined in Algorithm 2. Once the  procedure is finished we aim to have a  dataset comprised of stimuli, which are not associated with any individual more than what $\hat p(u)$ indicates. Note that for this extension, we need access to $D$ in Step 2, to calculate $(\hat p(u))_{u\in \mathbb U}$.  Additionally, this extension constitutes an important empirical addition for   AM-generation  to achieve anonymization for task $\mathbb X \rightarrow \mathbb U$. As this is not done in the original AM-generation described in Algorithm 1, the Algorithm does not explicitly attempt to preserve privacy for $\mathbb X \rightarrow \mathbb U$.

\subsection{Theoretical Intuition: Inhibiting a Strong Adversary}

We provide a brief analysis for the reasoning behind the stimuli inhibition. We use the   example from the previous subsection and assume an adversary, that wants  to identify Tom in an image $x\in \mathbb X$.  We define the adversary as an index function  $A: \mathbb X\rightarrow \{0,1\}$. This means that $A$  labels as 1 regions of $\mathbb X$ for positive identification, i.e., $A$ thinks it has identified Tom in an image, and 0 for negative, i.e., $A$ thinks the image does not include Tom. 
$A$ only needs access to a single datapoint (in our case image $x$) in order to perform the attack, i.e., to identify (or not identify) Tom in the image, i.e., the datapoint.  As such, the proposed attack  differs from other attacks commonly used in related work  like the membership inference attack \shortcite{shokri2017membership} \footnote{given a datapoint and black-box access to a model the goal of membership inference attack is to determine whether the datapoint is in the training set used to train the model}. This happens because  the proposed attack  does not  extract information from a model, but instead it is performed directly on the data. In our example $A$, is a function with  predefined constraints, and it can also be a ML model  trained on (other) images of Tom to perform generalization on image $x$.

We want to quantify the success of $A$. To do this, we can define a score  to measure the success of the adversarial decision $A(x)$ given a point $x$.
A  simple way to score $A(x)$ is to assign  a positive point if $A$ successfully identifies $x$. Given the outcome  $X=x$,   if $A(x)=1$ and the outcome that label $U$ corresponds to Tom, then we award $A$ a point. We also award a point if, given $x$, $A(x)=0$ and  $U$  does not correspond to Tom. Otherwise, we award 0 points.     As such, we define a score function $S_A(U,X)$ as follows: if $A$ is correct $S_A=1$, else $S_A=0$. Formally, $\forall x\in \mathbb X$:

\begin{equation}
\small
S_A(U,X=x)=\begin{cases}
1, (A(x)=1\cap U=u_t)\cup(A(x)=0\cap U\neq u_t)\\
0,  else
\end{cases}
\end{equation}

where we  assume that $u_{t}$ is the class that corresponds to Tom in $\mathbb U$. For any given $x$, the conditional expectation of the score $\mathbb E_{U\sim p(u|x)}[S_A(U,x)|x]$ poses a quantifiable metric of the expected adversarial success for $x$. We define it as follows\footnote{ Notice that since $P(X=x)=0$, we need to specify the limiting procedure that produces $X=x$. To do that, we define a region  in  $\mathbb X$ for which any point is arbitrarily close to $x$ given a distance metric, $H^\epsilon_x=\{x':||x'-x||<\epsilon\}$. We have that for all $\epsilon, P(H^\epsilon_x)>0$. We define $\mathbb E[S_A(U,x)|x]=\lim_{\epsilon \to 0} \mathbb E[S_A(U,X)|X=x'\in H^e_x]$.}:

\begin{equation}
\small
\mathbb E_{U\sim p(u|x)}[S_A(U,x)|X=x]=\begin{cases}
1\cdot p(u_t|x) +0\cdot (1-p(u_t|x)), A(x)=1\\
1\cdot (1-p(u_t|x))+0\cdot p(u_t|x),  A(x)=0
\end{cases}
\end{equation}

Next, we separate $\mathbb X$ into three disjoint regions as follows: (1) $R=\{x|p(u_t|x)=0.5\}$, (2)  $R_+=\{x|p(u_t|x)>0.5\}$ and (3)  $R_-=\{x|p(u_t|x)<0.5\}$.  In our analysis, we assume that $\mathbb X$ is a compact set. Furthermore, let us define any adversary $A^*$ which is 1 in $R_+$ and 0 in $R_-$. For notational convenience, we will depict as $\mathbf{A}^*$  the set of all adversaries that satisfy this condition, i.e., $\mathbf{A}^*=\{A| (A(x)=1,x\in R_+)\cap (A(x)=0,x\in R_-)\}$. Then for any $A^*\in \mathbf{A}^*$  and for regions $R,R_+$ and $R_-$ the following hold:

\begin{theorem}
Given score $S_A$:  a) For any $A^*\in \mathbf{A}^*$, $\forall A',\forall x\in\mathbb X$,   $\mathbb E[S_{A^*}(U,x)|x]\geq$ $\mathbb E[S_{A'}(U,x)|x]$,   
b) $\forall x \in R$, given  any adversary $A$:  $\mathbb E[S_A(U,x)|x]=0.5$, and 
c) $\forall x \not\in R$, for any $A^*\in \mathbf{A}^*$:  $\mathbb E[S_{A^*}(U,x)|x]>0.5$
\end{theorem}

The proofs are included in Appendix D. However, the simple score function $S_A$ needs to be refined, because it assumes equal score despite relative distribution mass, which is problematic. For example, given $S_A$, if Tom is  under-represented in $p(u)$  and the maximum reachable conditional probability $max_{x\in\mathbb X} p(u_t|x) = 0.4<0.5$, then $\mathbf{A}^*$ ``collapses'' into a single adversary  $A^*(x)= 0,\forall x\in\mathbb X$. This means that in this case, any optimal $A^*\in \mathbf{A}^*$  would not ``see'' Tom  in $\mathbb X$. However, this is something we do not want in our analysis. We want to inhibit the  give-away features of all individuals equally during the formation of our data, despite potentially weak representation in the original data. Therefore,
we assign in the refined score function the value  $1/p$ instead of 1, where 
 $p$ is the total probability mass of the values of $U$ that corresponds to the particular correct case for $A$. If $A$ is not correct, we assign 0. As such, the new score function is: 

\begin{equation}
 S^r_A(U,X=x)=\begin{cases}
1/p(u_t), (A(x)=1)\cap (U=u_t)\\
1/(1-p(u_t)), (A(x)=0)\cap (U\neq u_t)\\
0, else
\end{cases}
\end{equation}

 where we trivially assume that $0<p(u_t)<1$. With this change,  the conditional expectation of the score becomes for $A$ and $\forall x \in \mathbb X$:

\begin{equation}
\mathbb E_{U\sim p(u|x)}[S^r_{A}(U,x)|x]=\begin{cases}
\frac{p(u_t|x)}{p(u_t)} , A(x)=1\\
\frac{1-p(u_t|x)}{1-p(u_t)}, A(x)=0
\end{cases}
\end{equation}

This conditional expectation is recalibrated equally for both cases of a correct decision, regardless of  the adversary. As before, we separate $\mathbb X$ into three regions: (1) $R^r=\{x|p(u_t|x)=p(u_t)\}$, (2)  $R^r_+=\{x|p(u_t|x)>p(u_t)\}$ and (3) $R^r_-=\{x|p(u_t|x)<p(u_t)\}$. Furthermore, we define as above the set $\mathbf{A^r}^*$ of all adversaries ${A^r}^*$ which are 1 in $R^r_+$ and 0 in $R^r_-$, i.e., $\mathbf{A^r}^*=\{A| (A(x)=1,x\in R^r_+)\cap (A(x)=0,x\in R^r_-)\}$. Then for any ${A^r}^*\in \mathbf{A^r}^*$, and regions $R^r, R^r_+$  and $R^r_-$ similarly as Lemma 1 we have:

\begin{theorem}
Given score $S_A^r$:
a) For any ${A^r}^*\in \mathbf{A^r}^*$, $\forall A',\forall x\in\mathbb X$,    $\mathbb E[S_{{A^r}^*}(U,x)|x]\geq \mathbb E[S_{A'}(U,x)|x]$,
b)  $\forall x \in R^r$, given  any adversary $A$:  $\mathbb E[S^r_A(U,x)|x]=1$, and
c) $\forall x \not\in R^r$, for any ${A^r}^*\in \mathbf{A^r}^*$:  $\mathbb E[S^r_{{A^r}^*}(U,x)|x]>1$
\end{theorem}

Now, any optimal adversary  ${A^r}^*\in\mathbf {A^r}^*$ can in our example identify Tom when $p(u_t|x)>p(u_t)$. Based on these results, given the score $S^r_A$, region $R^r$ guarantees a low expected score for any $x \in R^r$. This  is not the case for the rest of $\mathbb X$. As such, given that our generating process can generate useful data $x\in R^r$, we have for each of these datapoints a guarantee about their inhibitory value  regarding any adversary $A$.   In practice we attempt to reach the intersection of the respective $R^r$ regions for all individuals in the data with the minimization of the cross entropy between $h_{TU}$ and $(\hat p(u))_{u\in \mathbb U}$, i.e.,  $L_U$, which approximate  distributions $(p(u|x))_{u\in \mathbb U}$ and $(p(u))_{u\in\mathbb U}$ respectively. Even though this intersection can be the empty set, by minimizing this loss we aim to ``blunt'' any strong class belief for $U|X$. Specifically, if we assume continuity in  $\mathbb X$ for $p(u=u_c|x)\forall u_c\in \mathbb U$, and we manage to find $x\in \mathbb X$ such that $L_U(x)<T_U$ for small enough value $T_U$, we push all scores of the optimal adversaries of all $u_c\in \mathbb U$ to be small at the same time. As such, we aim to generate $x_s\in \mathbb X$ which  achieve a good proximity compromise across the $R^r$ regions of the different individuals, in $\mathbb U$, i.e., $h_{TU}(x_s)$ is close to $(\hat p(u))_{u\in \mathbb U}$. Note that in our analysis, we assume  that $h_{TU}$ is a continuously differentiable function (almost) everywhere in $\mathbb X$. 

In summary: the above analysis finds, based on a score function, regions where all  adversaries have weak identification potential. Then, with the inhibition described previously we attempt to generate data from these regions. It is important to note that there is no theoretical guarantee that this  attempt will succeed,  because of three reasons. First, there is no guarantee that a common region, i.e., the intersection of all different $R^r$ for all classes of $U$, exists. Second,  all functions used during the proposed approach are approximators of the true functions whose conditions define $R^r$. Third, even if the functions used were the true functions, and the appropriate intersection existed, there is no guarantee that the optimization  of $G_{AM}$ will converge. However, despite these considerations, in all our experiments, the process empirically yielded promising results.

\subsection{Why Does AM Work?}

An important question regarding the AM stimuli is why would $D_S$ provide an efficient way to learn for another (similar) classifier $h_S$. For example, why would using random noise of the same dimensionality as the input space, with $h_T$ labelling each noise sample, be inferior than using a stimuli dataset. Indeed, preliminary experiments on MNIST show that training with 1000 stimuli yields  much better performance on new data than training with   1000 noise samples \footnote{Accuracy on MNIST test set: Random uniform noise: 0.20, AM-Stimuli: 92.91.}. In both cases $h_T$ labelled the samples. We hypothesize that this characteristic can be
explained by the fact that AM-stimuli localize ``important'' regions of the input space. If $\forall x\in \mathbb X$, $h_T(x)$ provides a good approximation to $(p(y|x))_{y\in\mathbb Y}$, then, AM stimuli, are  by definition low-entropy (i.e., high belief) points for $(p(y|x))_{y\in \mathbb Y}$. We assume that a diverse low-entropy dataset can provide a more efficient way to learn than  random noise since  important information about the features of each class is more efficiently captured. For this to hold and to enable proper generalization  we need: (1) that $h_T$ properly approximates $(p(y|x))_{y\in \mathbb Y}$ in $\mathbb X$, (2) that the stimuli are diverse in the sense that they capture all important low-entropy regions per class, (3) that $h_S$ is able to learn and generalize despite the differences  between the stimuli and the true data distribution. We discuss the last point in more detail in Section \ref{OtherCl}.    

\subsection{A Working Example: Creating $D_S$}

We  use the example from Section 2.2.4 to further showcase how the method works. Assume that we have a dataset $D=\{(x_0,y_0)...(x_N,y_N)\}$ for the task of identifying which face is smiling. In our example, $\{x_i\}_{i=1}^N$ are RGB images of a specified size that include a face and $\mathbb Y=\{0,1\}$ is a binary decision    whether the face is smiling or not. John, Mary, and Tom ($\mathbb U=\{0,1,2\}$) contributed to  $D$ a total  of 150 datapoints, i.e.,   $|D|=150$. Assuming that  we know  for each image the individual it  comes from, we train two classifiers: (1) $h_T$ learns $\mathbb X\rightarrow \mathbb Y$,  to distinguish between faces which are smiling or not, and (2) $h_{TU}$ that learns $\mathbb X\rightarrow \mathbb U$,  to distinguish between faces that originate from John, Mary, and Tom. We  apply Algorithm 2, and  repeatedly maximize  activations  of the class neurons of $h_T$. During the same optimization step, we inhibit the class activations of $h_{TU}$  based on $\hat p(u)$ so that there is no strong association towards any individual in the stimuli of $D_S$. If $h_T$ and $h_{TU}$ generalize reliably, and the procedure is tuned appropriately, $D_S$ will contain useful  information for  $\mathbb X\rightarrow \mathbb Y$, while withholding similar information about $\mathbb X\rightarrow  \mathbb U$.  The intuition behind the second part of this hypothesis, is that under these conditions, any given point $x_s\in D_S$, will not contain more probability mass than the expected mass for any value of $U$, independent of $x_s$.

\subsection{Choices for Important Hyper-Parameters}

In this subsection, we describe our choices for hyper-parameters for AM generation.

\subsubsection{Thresholds $T_Y$ and $T_U$}

Specifying appropriate values for the two thresholds is very important for the proper function of the AM-Generation. To do this, we use the following simple empirical criterion: We set the $T_Y$ per class, equal to the median activation  of the set of datapoints from $D$ which belong to each class. This is a natural criterion which mimics the activations of the real data for $h_T$. If this threshold is too high, it could result in  slow data synthesis because the objective becomes too hard, and even in low performance in some cases. If we notice such a behaviour, we can make the acceptance threshold $T_Y$ lower (e.g., for a certain class equal to the minimum activation of all datapoints belonging to this class).

 Identifying appropriate values for $T_U$ is less trivial. An initial idea is to specify a threshold that is based on random noise. For example, we could generate random noise inputs, calculate the mean loss for these inputs, and use this number as a threshold $T_U$.  The random noise  input $x_{z_i}$ can be drawn  from a uniform distribution with dimensionality and bounds equal to that of the input space. Then we calculate $T_U$ as: 

\begin{equation}
    T_U= -\frac{1}{N_z} \sum_{i=1}^{N_z}\sum_{u_c}^{N_p-1}\hat p(u_c) \cdot \log h_{TU}^{(u_c)}(x_{z_i})
\end{equation}

 with $N_z$ the number of random noise samples. This idea is based on the fact that a random noise sample would  be very unlikely to contain any identifying features for a specific individual $u_c\in \mathbb U$, and thus its activation of $h_{TU}$ will be close to $\hat p(u_c)$.   Empirically, this idea yields consistent results for  data of high dimensionality with clear class separation.  However, for data with relatively low dimesionality and with less trivial class separations, there were cases in which the loss for the noise samples was not  smaller than that  of the true data. The majority of medical  tasks we investigate fall into this case. As such, in these cases, we use a simpler strategy, and utilize a $T_U$ value  which is able to satisfy two requirements: (1) it is substantially smaller than the smallest datapoint loss in $D$, i.e., $T_U<< min_{x\in D} \{-(\hat p(u_c))_{u_c}\cdot  (\log h_{TU}^{(c)}(x))_c^T\}$ and (2) it is reachable by the optimization in Algorithm 2. 

\subsubsection{Mini-Batch Sizes $m$ and $m'$}

Empirically, the size of the training mini-batch $m$ does not play a very crucial role in the success of the AM-Generation. Larger mini-batch sizes generally lead to smaller amount of steps until convergence for the transformation $G_{AM}$  at the expense of slower execution times per-step. 

We keep in all experiments the size of the candidate batch $m'$ smaller than $m$.
In all experiments, $G_{AM}$ of the same AM loop tended to produce  similar outputs especially in the cases which are on the same optimization step. Therefore, we keep $m'$ very small and use values for $m$ from 8 to 64 and for $m'$ from 1 to 5. 

\subsubsection{Learning Rates $l_Y$ and $l_U$}

The learning rates $l_Y$ and $l_U$ play an important role in the success of the optimization of $G_{AM}$. As $G_{AM}$ has a dual optimization objective, we found that using smaller learning rates in relation to the ones that are used for $h_T$, helped the learning process. In most of our experiments, we use the same value for $l_Y$ and $l_U$, i.e., values from 0.0001 to 0.001. 

\subsubsection{Optimizers }

Any gradient optimization algorithm can be used to train $G_{AM}$. We used the Adam optimizer \shortcite{kingma2014adam} in all of our experiments.

\section{Tasks and  Datasets}
\label{sec:Datasets}
The previous section presents a new approach to use modified noise, transformed to stimuli instead of the real data as a means of teaching new classifiers about the class separation for a task we are interested in. Furthermore, this approach can provide protection against potential adversarial information leakage attacks. These two advantages of the algorithm are individually evaluated in Section 5 and 6. Both evaluations are  mainly focused on   the application scenario of sleep apnea detection. Therefore, we describe in this section the application scenarios and the datasets, before we (1) investigate in Section 5 how well different ML methods or architectures can generalize when trained with stimuli generated from the knowledge of $h_T$,  based on Algorithm 1 from Section 2,  and (2) investigate in Section 6 whether using stimuli instead of real data can provide protection for certain adversarial attacks.

We want to detect Obstructive Sleep Apnea (OSA) with low-cost sensors and using ML-based analysis on smart-devices. Sleep Apnea events are defined as the cessation of airflow for at least 10 seconds or reduced airflow by at least 30\% (American Academy of Sleep Medicine - AASM, \shortcite{ApneaAASM}). 

OSA can be diagnosed by Polysomnography(PSG) in a sleep laboratory or by polygraphy at home.
A variety of physiological signals are commonly used  for OSA diagnosis, including  electrocardiogram (ECG), electroencephalogram (EEG), electromyogram (EMG), electrooculograph (EOG), oxygen saturation,
heart rate, blood pressure and respiration from the abdomen,
chest and nose. To detect sleep apnea  a sleep expert   visually inspects the graphical representation of the time-series data to detect and label apneic events.

 OSA is a very common, yet severely under-diagnosed
disorder. In Norway, it is estimated that around 25\% of all middle-aged Norwegians are at high risk of having OSA, yet approximately 70–80\% of all cases are expected to be undiagnosed \shortcite{punjabi2008epidemiology}.  In our work, we use data from a large clinical study, called A3 study, at the Oslo University Hospital and St. Olavs University Hospital, Trondheim \shortcite{traaen2019treatment,traaen2020prevalence}. In this study, sleep monitoring data from several hundred patients is collected and analyzed. As such, this is data that is collected every day in clinical settings  to address a severe health issue. However, privacy regulations do not permit the sharing of the data and prevent reproducibility of the results in this paper gained  with the A3 data. Therefore, we use in addition to the A3 data the well-known open sleep monitoring data set called Apnea-ECG  \shortcite{ApneaEcg}.

Furthermore, even though our main focus is sleep, we do not want to focus  only on one medical condition or task,  to establish a broader understanding of the capabilities of AM generation. Thus, to investigate different signal types and  healthcare tasks, we use the open Sleep EDF dataset \shortcite{867928} for sleep stage classification, and  evaluate our approach with the use the EEG signal. Sleep stage classification is a different task from sleep apnea detection, that includes the following classes (stages):  W (awake), 1 (transitional phase), 2 (light sleep), 3\&4 (deep sleep), R (REM sleep). The goal is to identify from patterns of the sensory input the stage of the individual in the particular time window.
Additionally, the use of the Sleep EDF dataset gives us the opportunity to include more individuals in our experiments while using an open-access dataset.

For most humans, sleep monitoring data are  hard to evaluate and can contain noise and artifacts. Thus, we start our investigation with a  simpler problem and  a well understood dataset,  namely digit recognition with the MNIST dataset. The insights gained with MNIST help to properly experiment and interpret  results with the more challenging sleep monitoring datasets. Furthermore, the use of multiple different types of data indicates the generalizability of the proposed approach. 

In total, to evaluate our approach  we use four  datasets which correspond to three different tasks, namely digit classification, sleep apnea detection, and sleep stage classification:

\begin{itemize}
    \item  \textbf{MNIST} is a well-known database  containing a training set of 60000 28$\times$28 black and white images of 0-9 handwritten digits and a test set of 10000 digits. Out of the 60000 training data we use 5000 as validation set.
    
    \item  
     \textbf{Apnea-ECG} is a well-known open database from Physionet  which contains   data from multiple sensors capturing ECG, respiration from the chest and abdomen, nasal airflow (NAF), and oxygen saturation. The data  contains sleep recordings  from 70 patients with durations  of 7- 8 hours.  Apnea-ECG   has been collected in a sleep laboratory with PSG and preprocessed such that it contains only high quality data. From Apnea-ECG we mainly use  the NAF signal, because it can  be used adequately to  train a classifier to recognize apneas and yields the best single signal performance
 among all the respiratory signals as shown in our previous work \shortcite{kristiansen2018data} . We use the eight sleep recordings that contain the NAF signal  (i.e., a01, a02, a03, a04, c01, c02, c03, b01). The sampling frequency of the sensors is 100 Hz  and all recordings contain labels for every minute window of breathing that signifies whether the minute is apneic or not, i.e., whether the person is experiencing an apneic event during this minute or not.  
 
 \item  The \textbf{A3 study} \shortcite{traaen2019treatment,traaen2020prevalence} investigates the prevalence, characteristics, risk  factors and type of sleep apnea in patients with paroxysmal atrial  fibrillation. The data were obtained with the use of the Nox T3 sleep monitor \shortcite{Nox} with mobile sleep monitoring at home, which in turn  results into lower data quality than data from PSG in  sleep-laboratories. An experienced sleep specialist scored the recordings manually using the Noxturnal software \shortcite{Nox} such that the beginning  and end of all types of apnea events is marked in the time-series  data. To utilize the data for the experiments in this paper, we labeled  every 60 second window of the data as apneic (if an apneic event  happened during this time window) or as non-apneic. The data we use in  the experiments is from 438 patients and comprises 241350 minutes of  sleep monitoring data. The ratio of apneic to non-apneic windows is  0.238. We utilize  the NAF signal from the A3 data in the experiments,  i.e., the same signal we use from Apnea-ECG .

 \item The \textbf{Sleep-EDF Dataset} \shortcite{867928} contains 197 whole night PSG sleep recordings including  EEG,  EOG,  chin EMG, and event markers. The data are separated into 30 second periods, and each period is given a label which corresponds to a sleep stage, as described above,  movement (``M''), or not scored (``?'').  From this dataset we utilize a different signal than  the ones used for the  OSA detection experiments, namely EEG.  We base our feature extraction and feature engineering scripts on the  \textit{mne Sleep Stage classification from PSG data} scripts for python \shortcite{mne,chambon2018deep}. As indicated from these sources, we combine sleep stages 3 and 4 into one class and we exclude non-scored and M segments. As such we use 5 classes.
\end{itemize}

For all experiments, we preprocess the data. Specifically, for MNIST,  we  rescale the data from 0-255 to 0-1. We  downsample  Apnea-ECG and A3 to 1Hz. Finally, for   the Sleep EDF dataset,  we perform feature engineering  and  extract  EEG features based on relative power in specific frequency bands, with the use of Welch's method for spectral density estimation  \shortcite{welch1967use}. Two EEG channels are used. Additionally, we increase the sampling frequency per power band.  Specifically, instead of the 5 commonly used bands (i.e., gamma, beta, alpha, theta, delta), we  use 30 bands (one per 1Hz).  We do this to gather more information per power spectrum, and since preliminary experiments showed increased performance for the sleep stage classification task.

\section{Preliminary Empirical Analysis}
\label{sec:prelim}

Before we proceed with the generalization and anonymity experiments, we include in this section a preliminary analysis about important characteristics of the generated stimuli. It is important to note that for all experiments and figures of this section, no inhibition is performed (i.e.,  Algorithm 1 is used).

 \subsection{Stimuli Visual Appearance}
 
 \par Figures \ref{fig:AM_REAL_Data} and \ref{fig:Digit_Synth} show examples of AM stimuli from MNIST and Apnea-ECG. Although the stimuli are as expected  not realistic, they potentially  contain implicit information about their respective classes (see Section 5). As such, a classifier that learns from the stimuli is able to a certain extent to generalize on the real data, depending on the  algorithm used (see Section \ref{OtherCl}). In both cases, the  stimuli can be drastically different, even when they represent the same class. This diversity is beneficial for  the student's learning. Additionally, it is hard in both cases to distinguish  between the classes for the stimuli.

To  verify that the synthetic data did not contain real datapoints from the original training data, we found the  closest distance datapoints between the AM produced dataset and the true dataset for all the datasets we experimented with (MNIST, Apnea-ECG, A3). In  Appendix C we present the closest neighbors for the datasets we used. In all cases, no datapoint from the training set was visually similar to any of the datapoints in the synthetic data.

\subsection{Stimuli Feature Maps}

 To investigate the  learned features of $h_S$, we train a CNN model with a two convolutional-layer architecture. The model is trained on MNIST stimuli. Specifically, the trained model has an identical architecture to  the $ID$ architecture used later in our experiments, which is described in Appendix A.  Figure \ref{fig:FeatVis} shows for both layers the feature maps of the model for  both a datapoint from the real data of label 2, and   a stimulus of the same label.  For our implementation we followed the tutorial by J. Brownlee  on filter and feature map visualization \citeyear{featvis}.

\begin{figure}[h]
    \centering
    \includegraphics[scale=0.35]{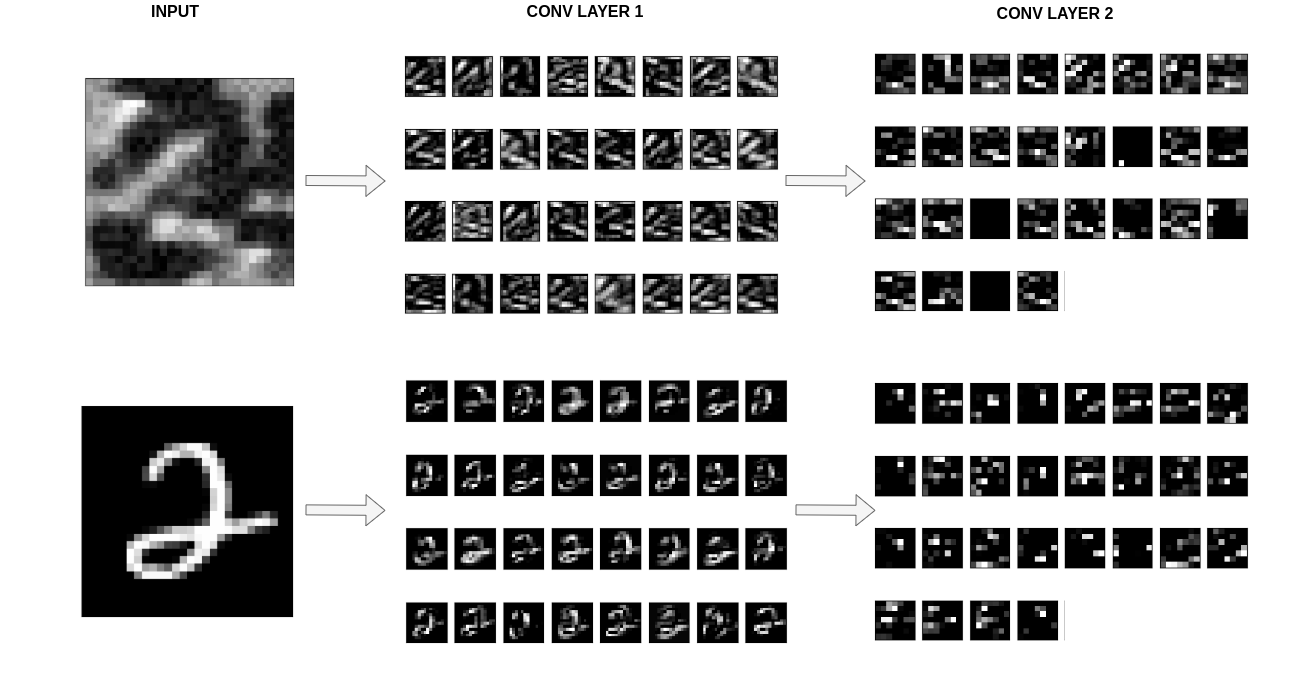}
    \caption{Feature maps for $h_S$ (CNN $ID$ architecture). }
    \label{fig:FeatVis}
\end{figure}

In both cases, the feature maps closer to the input of the model capture  fine details of the respective datapoint. As we progress into the second layer, the feature maps show less  details.
This pattern is to be expected, as the model abstracts the features from the image into more general concepts in order to make a classification. 

In both cases, in the first layer the model identifies patterns that could generally be associated with the digit 2. Intriguingly, we observe that in the case of the stimulus most maps of this layer seem to indeed prioritize patterns from the input stimulus that resemble the proper digit.  The second layer feature maps are in both cases not as easily interpretable as the maps from the first layer. However, high level patterns that could aid towards the classification of the digit as 2 emerge for both cases. Interestingly, for the case of the stimulus, we notice that some of the second layer maps seem not  to contain representational information. 

\subsection{Stimuli Spread on the Input Space}

We hypothesize that, as each stimulus includes implicit class information, the stimuli dataset would need to be more ``spread out'' in the input space relative to the real dataset. This way the stimuli could potentially constitute a more diverse sample which could compensate for the inability of each datapoint to include direct class information.

To measure the ``spread'' of each class of the stimuli dataset we use a simple empirical measurement, i.e., the median maximum euclidean distance between the stimuli of the class, on the MNIST dataset. As expected, the stimuli have higher median maximum distances across all classes when compared to the real data. Specifically, the stimuli yielded an average value of 18.03, whereas the real data a value of 11.78.   In Appendix F, we include the detailed values per-class, and  we additionally perform an ablation study where we remove the randomness operators mentioned in Section \ref{Method}.

\section{Generalization  and  Customization}
\label{sec:genandcust}
We evaluate the generalization capability of the proposed approach  when $h_S$ is either similar or dissimilar to $h_T$,  and compare it with  state-of-the-art methods. To demonstrate that data from AM-based stimuli generation generalize successfully, we use Algorithm 1 to generate data. In this section, we focus on  MNIST and the closed A3 dataset, and include Apnea-ECG for completion.

\subsection{Experimental Set-up}
\label{Setup}

 We select the VAE and gradient-penalty Wasserstein GAN (WGAN) \shortcite{pmlr-v70-arjovsky17a,gulrajani2017improved} for comparison    since both frameworks are very well-established (in terms of citations and github repository projects).  VAE is a more basic  non-adversarial generative model, and serves as an initial comparison basis. WGAN is  a more modern, general variation of the very successful GAN framework. The WGAN is especially stable during training and exceptionally good at avoiding mode-collapse. Furthermore, it is  able to produce very realistic and stable results for our tasks.

\begin{figure}[h]

\begin{subfigure}{0.5\textwidth}
\centering
\includegraphics[width=7cm,height=2cm]{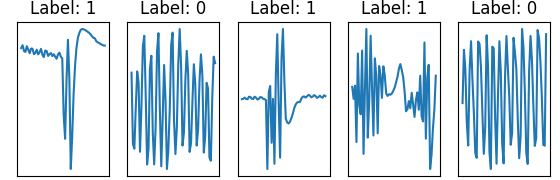}
\caption{Apnea-ECG, Real Data}
\end{subfigure}
\begin{subfigure}{0.5\textwidth}
\centering
\includegraphics[width=7cm,height=2cm]{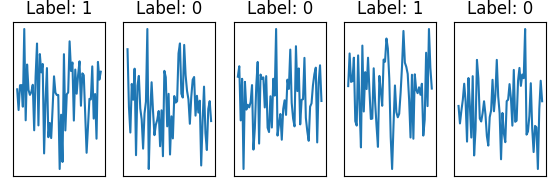}
\caption{Apnea-ECG, Stimuli}
\end{subfigure}
\caption{Real data (a)  and stimuli (b)) for  NAF signal   Apnea-ECG. The labels of the datapoints are shown above the images of the data. The y-axis of Apnea-ECG graphs corresponds to mV and x-axis to seconds (windows of 60 seconds). In all cases, the minimum required threshold used to generate AM-stimuli was $T_Y>0.96$. }
\label{fig:AM_REAL_Data}
\end{figure}

  In all experiments we perform four steps: (1) train $h_T$, (2)  use AM, WGAN or VAE to generate a synthetic dataset  (3)  train   $h_S$ with the synthetic data, and (4)  evaluate $h_S$  with the  test set.  For MNIST we use the original test set. For Apnea-ECG and A3, we create a test set  by randomly sampling from the datasets. We use 15\% of the   Apnea-ECG data and 20\%  of the A3 data as test set. We choose to perform random sub-sampling (i.e., Monte Carlo) cross validation as our validation method  over K-fold cross validation since we wanted to (1)  use large enough test sets, and (2) repeat the experiments enough times so that we can get reliable statistics, while maintaining a large enough dataset. Based on these requirements, using Monte Carlo cross validation offers a better  solution than cross validation, since we can more freely control these parameters (i.e., test set size and number of repeats of the experiements). For the  WGAN model,  we use a conditional WGAN proposed by Mirza \& Osindero \citeyear{mirza2014conditional}. For VAE we use a conditional VAE, based on the implementation by  Kristadl \citeyear{cVAE}.

\begin{figure}[h]

\begin{subfigure}{\textwidth}
\centering
\includegraphics[scale=0.8]{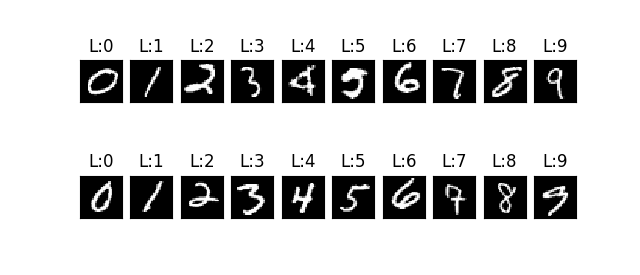}
\caption{Real Digits}
\centering
\includegraphics[scale=0.8]{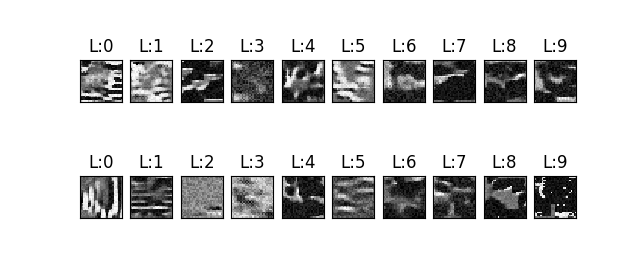}
\caption{Stimuli Digits}
\end{subfigure}
\caption{Real data (a) and stimuli (b) for   MNIST  digits. The labels of the datapoints are shown above the images of the data (L:).  In all cases, the minimum required threshold used to generate AM-stimuli was $T_Y>0.96$. }
\label{fig:Digit_Synth}
\end{figure}

Different  configurations for the AM-, WGAN-, VAE-based set-ups  can yield  very different results. Therefore, we perform in a pre-study Steps (1) to (4) with various hyperparameter configurations. We then choose the configuration that yields the best performance of $h_S$.  With the configuration chosen from the pre-study we repeat Steps (1)-(4) to perform the actual experiment. With the selected  configurations, we train $h_T$ in Step (1), and  we  create the synthetic dataset in Step (2). With this dataset we evaluate the performance of $h_S$ on the  test set by  training and testing  $h_S$  on Steps (3,4). We use a validation set and  save periodically  the parameters of $h_S$. We choose the parameters which yield the best validation performance and use this model to classify the test set.

For the WGAN and VAE comparisons, we assume that the host  releases synthetic data, which are   synthesized from a  WGAN generator or a VAE decoder, in order to  train  $h_S$.   For this comparison to be on equal terms, we design the generative models such that the WGAN generator ($G_{WGAN}$) and the VAE decoder ($D_{VAE}$) have a similar architectural configuration as $h_T$, since $h_T,G_{WGAN},D_{VAE}$, are the  models that contain the task knowledge in the respective methods.  Furthermore, we  train $h_S$ with the original training data and evaluate it on the test set  in order to understand the impact of   similarity (in terms of architecture and size) onto the performance of $h_S$.

 \par We investigate five different $h_S$  architectures, called ID, S, VS, L, and LL.   ID  is identical to  $h_T$\footnote{ The network $h_T$ corresponds to the TEACHER of Figure 1.}. S  is  similar to $h_T$,  but with  half the number of weights per layer.  VS   has a quarter of the weights per layer compared to $h_T$\footnote{With the exception of MNIST, where we experiment with an even smaller design. For more details please refer to Appendix A.}.  The architecture of L is similar to $h_T$, but without  the second to last fully connected layer.  In LL,  the second and third to last layers (either fully connected or convolutional) are removed.   For 
 all experiments, we aim for the WGAN generator and the VAE decoder to have a similar number of parameters and architecture to $h_T$. For the AM case we use as labels the softmax outputs of the output neurons (soft labels). For all  datasets we investigated, we did not achieve significant performance gains by utilizing larger architectures than the ones we chose as $h_T$. Furthermore, we focus our evaluation on smaller classifiers potentially for use in a resource constrained environment. All architectures for all experiments of this section are presented in Appendix A. For the MNIST and A3 datasets we also present the state-of-the art generalization performances on the real data, to give an indication of the best achievable performance on each investigated dataset.  For Apnea-ECG, to the best of our knowledge, no related work exist that uses specifically the same signals, subjects, and separations that we perform in this work. As such we   do not mention these numbers for Apnea-ECG since they are not sufficiently related in our particular experimental context.

\begin{figure}

\begin{subfigure}{0.51\textwidth}
\includegraphics[width=\linewidth]{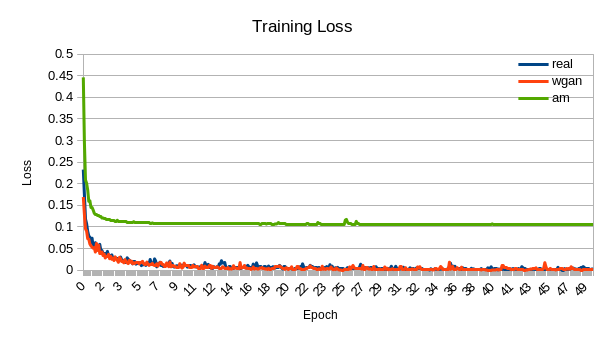}
\caption{Training Loss} \label{fig:1b}
\end{subfigure}
\hspace*{\fill}
\begin{subfigure}{0.51\textwidth}
\includegraphics[width=\linewidth]{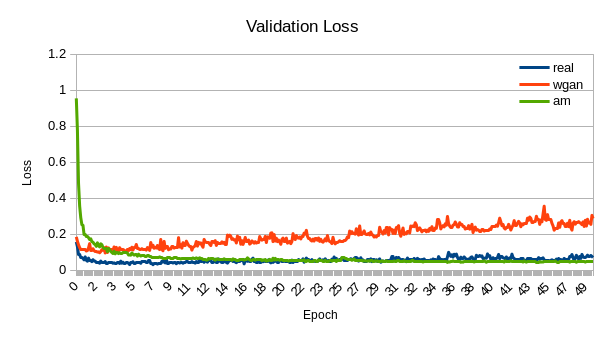}
\caption{Validation Loss} \label{fig:1b}
\end{subfigure}
\caption{Train (a) and Validation (b) loss for the real dataset (blue), WGAN (orange) and AM generation (green) with MNIST.} \label{fig:Loss}
\end{figure}

\subsection{Digit Classification}

We follow the aforementioned set-up and investigate the performance of $h_S$. As $h_T$ we use  a convolutional neural network  loosely based on LeNet-5 \shortcite{lecun1998gradient}, but with one  convolutional layer less and more weights per layer. Based on this, $h_T$ comprises   two convolutional, one max-pool  and two  fully connected layers (conv2D, maxpool, conv2D, fc1, fc2,  dropout on fc1). All layers include rectifier linear unit (relu) activations. We choose this architecture since it is a well-established model for digit classification. Furthermore,  we use more weights per layer to have a larger initial model as $h_T$, since it also achieves better performance  on average  than the thinner variants we experimented with. Similarly, the WGAN generator and the VAE decoder are  deconvolutional  networks with two fully connected and two deconvolutional layers with a  few more parameters per layer than $h_T$. ID has  the same architecture as $h_T$. S, VS,L, and LL  are structured as defined above. The third to last layer of LL is a convolutional layer. For more details on the different architectures, weights per layer, filter sizes, including the AM generator,  please refer to  Appendix A.  $h_T$  is trained with a batch size of 128 for 3125 batch iterations (see Section \ref{sec:Discussion1}). In this experiment, we generate 60000 AM stimuli as well as   WGAN and VAE  data. The experiment is repeated five times.

\begin{table}[h]
\small
\centering
\begin{tabular}{ | p{0.4cm}| p{1.5cm}|p{2.75cm}|p{1.6cm}|p{1.6cm}|p{1.6cm}|p{1.6cm}|}

\hline
&Structure&MaxLayer&Real Data &VAE& WGAN &AM\\
\hline
ID:&cmcmdd&c:32 - d:1372&99.20$\pm$0.02&91.69$\pm$0.64&97.63$\pm$0.08&98.64$\pm$0.24\\
S: &cmcmdd&c:16 - d:588&99.11$\pm$0.02&91.64$\pm$0.77&97.56$\pm$0.03& 98.19$\pm$0.06\\
VS: &cmcmdd&c:16 - d:64&98.73$\pm$0.13&89.97$\pm$0.65&97.24$\pm$0.06& 97.45$\pm$0.11\\
L: &cmcmd&c:32 - d:1372&98.89$\pm$0.04&88.80$\pm$0.91&97.50$\pm$0.05&98.04$\pm$0.05\\
LL: &cmmd&c:32 - d:1568&98.68$\pm$0.04&86.16$\pm$1.01&97.05$\pm$0.06&94.86$\pm$0.12 \\
\hline
\end{tabular}
\caption{Accuracy of $h_S$ for MNIST experiment, and structures of  $h_S$. Layer notation: 
c: convolutional layer, m: maxpooling, d: fully connected. The  MaxLayer column depicts the maximum number of channels and neurons in a single layer  used for  convolutional layers and  fully connected layers respectively. MNIST state-of-the-art performance (accuracy): 99.87, \shortcite{byerly2021routing}. }
 \label{tableMNIST}
\end{table}

Table \ref{tableMNIST} summarizes the results from the MNIST experiment.  AM outperforms the similar   WGAN and VAE models  for all $h_S$  architectures, except LL for the WGAN data. The differences between the AM,  WGAN and VAE  results are statistically significant for the one-tailed paired t-test ($p=5\%$) for all $h_S$ architectures, except  VS for the WGAN data. These differences could potentially be attributed to a better estimation of the class boundaries of the test set from the AM stimuli than from the WGAN or VAE data. Additionally, we hypothesize that $h_T$ contains more information about the classification than  the generative models, since its loss  is directly determined by the classification goal, which is not the case for the WGAN, or the VAE. Thus, if this knowledge is sufficiently mapped on a dataset, it could represent better the class separation boundaries than a dataset from the WGAN generator or the VAE decoder which potentially only focus on realistic  class data.   One interesting trend is that smaller architectures of $h_S$ lead to steeper performance drop for AM than for WGAN. This is a recurring phenomenon in all  experiments and we discuss it in detail in Sections \ref{sec:Insights} and \ref{OtherCl}.  The  performance of $h_S$ trained on  real data exceeds that of all other methods  in all cases,  as expected. 

\paragraph{Training and validation loss:}  We calculate the training and validation loss throughout the training of the $ID$ model  for the WGAN and AM datasets. We present the results in Figure \ref{fig:Loss}. The training loss of the AM dataset is  stable and larger  than the training loss for the Real  and WGAN  datasets. However, we notice for the AM case a  much better and more stable behavior for the Validation set.   We expect that  these effects can be attributed in part to the use of soft labels. We discuss these effects in more detail in Section \ref{sec:Insights}. Also, the validation loss of the WGAN increases  faster than  for the real dataset. This could be attributed to the lower variety of the WGAN data in relation to the real data.

\subsection{Apnea Detection}
For the apnea detection experiments, we repeat the same procedure as in the previous experiment. For Apnea-ECG, we use MLP instead of a CNN to investigate how  the  procedure changes with  a different architecture as $h_T$.  For the A3 dataset, we use a 1D convolutional neural network (see details in Table \ref{tableApnea} and Appendix A). All activations are relu, and dropout is used in both datasets for the second  to last fully connected layer. The generators and encoders of the WGAN and VAE for the two datasets correspond to similar sizes and architectures to  $h_T$. Similar architectural  conventions apply to the different $h_S$ we experimented with, based on the definition in Section \ref{Setup}. 
\par It is  important to mention that  we use for   apnea detection the kappa  coefficient \shortcite{cohen1960coefficient} as performance metric. We make this choice because the kappa coefficient better  captures  performance characteristics in a single metric than accuracy, as it takes into account the possibility of agreement by chance between the two annotators (real labels and predictions). Due to this property, it   succinctly and accurately reflects the performance of binary classifiers with imbalanced data, which is important for SA detection. Additionally, for completion we  include the accuracy, sensitivity, specificity and precision performance metrics in Appendix G. We use a  batch size of 128 and 1024 to train $h_T$ for Apnea-ECG and A3 respectively. With Apnea-ECG, we train $h_T$ for 100 epochs and with A3 for 5 epochs.  For Apnea-ECG we generate  with all generative models 5000 synthetic datapoints, whereas for A3 we generate 20000 synthetic datapoints. We repeat the A3 experiments 10 times and the Apnea-ECG experiments 20 times. We choose to repeat the experiments for more iterations than MNIST since, due to the use of random-subsampling on each iteration, our test set changes, thus imposing additional variance to our results. Furthermore, Apnea-ECG is a small dataset, and as such results are more varied when sub-sampling different test sets. This happens potentially because,  the amount of training datapoints is small, and as such a domain shift is ``naturally'' imposed between train and test datasets.

\begin{table}[h]
\centering
\footnotesize
\begin{tabular}{ | p{1.5cm}| p{1.6cm}|p{2cm}|p{1.5cm}|p{1.5cm}|p{1.5cm}|p{1.5cm}|}

\hline
&Structure&MaxLayer &Real Data & VAE&WGAN &AM  \\
\hline
AE ID:&dddd&d:360&88.30$\pm$0.46&49.55$\pm$1.9&79.87$\pm$0.76&85.49$\pm$0.42\\
AE S: &dddd&d:180&88.05$\pm$0.49&49.86$\pm$2.82&78.86$\pm$0.79&82.73$\pm$0.61\\
AE VS: &dddd&d:90&87.22$\pm$0.57&47$\pm$2.19&77.65$\pm$0.68&83.36$\pm$0.65\\
AE L: &ddd&d:360&86.58$\pm$0.53&51.12$\pm$2.17&79.26$\pm$0.67&83.70$\pm$0.53\\
AE LL:&dd&d:360&87.78$\pm$0.46&53.72$\pm$2.66&75.75$\pm$0.65&82.09$\pm$0.54\\
\hline
\hline
A3 ID:&cmcmcmdd&c:64 -d:512&61.64$\pm$0.33&\qquad -&54.73$\pm$0.48&58.82$\pm$0.45\\
A3 S: &cmcmcmdd&c:32-d:256&61.95$\pm$0.07&\qquad -&54.86$\pm$0.26&58.17$\pm$0.35\\
A3 VS: &cmcmcmdd&c:16-d:128&60.22$\pm$0.13&\qquad -&53.25$\pm$0.63&56.69$\pm$0.57\\
A3 L: &cmcmcmd&c:64-d:512&62.32$\pm$0.20&\qquad -&51.79$\pm$1.49&59.76$\pm$0.30\\ 
A3 LL:&cmcmmd&c:32-d:256&62.59$\pm$0.12&\qquad -&53.74$\pm$0.36&54.33$\pm$0.75\\ 
\hline
\end{tabular}
\caption{Kappa of the $h_S$ for Apnea-ECG and A3 experiments for different $h_S$ sizes when training with data generated from AM generation  and from conditional WGAN  and VAE of similar size. The MaxLayer column depicts the maximum number of neurons and channels of the net used. The VAE did not converge for the A3 Data. A3 state-of-the-art performance (kappa ($\times 100$), nasal-airflow signal):64.6, \shortcite{kristiansen2021machine} } 
 \label{tableApnea}
\end{table}
 Table \ref{tableApnea} summarizes the results for the apnea detection experiments. We observe that for all cases,  AM outperforms the WGAN and VAE. Contrary to all other cases, for LL the results are not significantly better for AM for the A3 dataset when compared to WGAN. For the one tailed t-test,  the p-value is: $0.24>0.05$. The performance drop of $h_S$ from  ID to   to LL is generally smaller for the WGAN case than for the AM case.  Additionally, for Apnea-ECG the performance of the L $h_S$ for the WGAN dataset is similar to the performance of the ID  WGAN $h_S$. Also notice the large performance difference between the models trained on WGAN data and AM data for  the Apnea-ECG dataset. We  attribute this difference in performance to the failure of the MLP WGAN generator to capture the important features of the dataset.
 
 Similarily, all VAE results are inferior to the AM generation cases for the Apnea-ECG dataset, and there are  much higher variations in  the VAE results. Furthermore, we were not able to reliably train the VAE model on the A3 data. Even though both reconstruction and distribution losses were decreasing, the formed dataset was not able to  train the student model. A potential explanation is the less distinct conditioning of the A3 data. 
 
The real data performance outperforms as expected the  WGAN, VAE and AM performance for all  cases. The real data performance is similar across the  $h_S$ architectures for both datasets. This  means that  the performance of sleep apnea detection seems less sensitive to  changes to the size of the network and the architecture than digit classification when trained with the real data. This could be due to the  less distinct class boundaries  in the case of sleep apnea detection relative to digit classification.

\subsection{Additional Insights}
\label{sec:Insights}

 In most  cases  we investigated,  the performance of $h_S$  is superior when training with AM stimuli than  when training with  WGAN or VAE generated data. However, as  mentioned above,  as we reduce the size of $h_S$, the
 drop in performance    is steeper with AM stimuli   than with WGAN data especially. One potential explanation of this phenomenon relates to an architectural mismatch between $h_S$ and $h_T$ (see Section \ref{OtherCl}). Another explanation relates to the type of labelling.  We hypothesize that the effect occurs due to  complex classification bounds of AM generated stimuli. These complex bounds are  potentially a result of the soft labelling. This hypothesis is generally strengthened by  the fact that the training loss during the training for the MNIST $ID$ experiment  is unstable and never goes to 0. This  means that even the largest network does not over-train to the AM data (see Figure \ref{fig:Loss}).  However, if we use hard labels, information of the beliefs of $h_T$ is being lost, and  we  get a steep performance drop among all $h_S$. As a result, soft labels are preferable.   This  relates to an additional property.  From Figure \ref{fig:Loss} we see that the validation loss for the AM data is exceptionally stable in relation to the real and WGAN data. This means that at least for this experiment, the generalization capability of the classifier trained on the AM data was not compromised by over-training. This indicates that  AM generation with soft labels introduces a form of regularization during the training of $h_S$.

\subsection{Generalization  with Different Classifiers as $h_S$}
\label{OtherCl}

An interesting point of the above results is the consistently high performance of $h_S$ with  similar architectures to that of  $h_T$. This raises the question whether it is possible to achieve a correspondingly high performance when using alternative architectures or classification methods as $h_S$.  We examined how a large dense deep net (DNN)  with relu activations, a RBF kernel SVM and a Random Forest ensemble with 50 trees perform when trained with AM stimuli  on the MNIST dataset. The AM performance  in all cases is   low. For  the SVM  we have: Real Data: 94.42, AM Stimuli: 39.96. For the DNN:   Real Data: 98.3, AM Stimuli: 81.06. For the RF: Real Data: 96.81, AM Stimuli: 30.82. This drop in relation to the real performance could  happen since the AM  stimuli are less correlated to the real test data than  the real training data is.  Especially for the SVM and the RF, for which the drop is much steeper than with the DNN, the difference in train and test marginal distributions can have a detrimental effect on their performance. This could happen since these models are not able to learn higher layer representations from the data, but instead they classify based on low-level decisions derived directly from  support vectors or comparisons of feature values. Another explanation relates to the use of the same architectures   for $h_T$ and $h_S$ in our experiments. For example,  it is known that the convolutional architecture  imposes  \textit{ inductive biases} over the learning algorithm \shortcite{battaglia2018relational}.   Thus, we suppose that when $h_S$ has  the same type of bias (i.e., the same architecture) as $h_T$, it can better capture and interpret how the stimuli correlate to the real data than another  algorithm without this prior  bias.

\section{Defenses and  Anonymity Properties}
\label{sec:defandanon}

In this section, we experimentally evaluate the protection against specific information leakage attacks  that $D_S$ or models trained on $D_S$ acquire.  Furthermore, we investigate additional potentially useful properties of $D_S$.

\subsection{Recording Association through Generalization}
We investigate an attack against anonymized open recordings.  This attack  is performed by recognizing,  through  generalization, associations among recordings that belong to the same individual.We perform two experiments to investigate the protection AM can offer against such an attack. To showcase a real world scenario we use  open datasets. First, we evaluate the feasibility of the proposed attack on a smaller scale, and  use  the respiratory sensor data of the open Apnea-ECG dataset. We show how the AM generated data can offer significantly better protection against identification in comparison to the  real data. Then, we further investigate the viability of AM on a different task, and   perform an additional experiment on a larger real-world open EEG dataset for sleep stage classification. We compare AM with differentially private versions of the generative models we experimented with, and show that AM constitutes an empirically superior alternative. In all experiments we investigate both Algorithms 1 and 2 for AM generation.

\subsubsection{Threat Model}
 The proposed attack draws inspiration from similar face identification tasks.
We assume that an adversary has access (potentially via a security breach) to sleep recordings from a group of individuals together with personal information about them like, e.g., their names.  Furthermore, we assume an open dataset which contains anonymized recordings (like Apnea-ECG), and that all recording data consist of  sensory signals. Furthermore, we assume that no common data between the open dataset and the breach exist. The goal of the attack is to probabilistically determine whether an individual has contributed a sleep recording to the open anonymized dataset.
 To do this, one can train a classifier to distinguish between  data of different individuals that belong to the breach, and  generalize to data from the open dataset. As such,  for example, in the sleep apnea scenario, the attack would assign minutes of data from the open dataset to individuals belonging in the breach. If the  data  sufficiently captures the idiosyncrasies of the different individuals, we  expect good classification performance on new data from the same person. We  ideally expect the following behavior: (1) for recordings of the open dataset that belong to individuals of the breach, we expect many minutes in the correct class,  i.e., classified correctly by the classifier as belonging to the correct individual; (2) for recordings from individuals that only have data in the breach, we ideally expect zero minutes assigned to the respective classes of the adversarial classifier; and  (3) for recordings from individuals that only have data in the open dataset, we expect the adversarial classifier to be more ``confused''. Thus, we would expect these windows to be more uniformly distributed among the classes of the classifier in relation to case (1). Note that an important requirement for this attack is that the recordings in the two datasets contain data from the same sensor types.

\begin{figure}[h!]
\centering
  \includegraphics[scale=0.5]{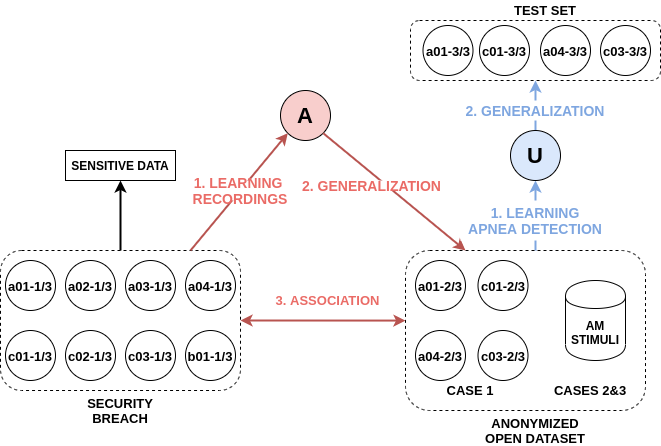}
\caption{Example showcase of the proposed scenario for the Apnea-ECG Dataset.  ``A'' depicts the adversary, which learns the sensory representations from the breach and generalizes on the open dataset. The adversarial goal is to identify which individuals from the breach have participated in the open dataset. A non-adversarial user ``U'' utilizes the open dataset for apnea detection and generalizes on a test set. As anonymized open dataset we  use the real raw data or AM stimuli. }
\label{fig:Attack}
\end{figure}

\subsubsection{ OSA Detection}

To simplify the experiment, we assume that unbeknownst to the adversary all individuals from the open dataset are included to the breach. Figure 
\ref{fig:Attack} illustrates the experiment. We use Apnea-ECG, and utilize all four respiratory sensors  for this task, because we need as much information as possible to map patterns to specific individuals. Apnea-ECG has eight respiratory recordings corresponding to different individuals. We partition each individual original recording into three disjoint  equal parts, and create the ``breach recordings'', the ``open dataset recordings'' and the test set in which we evaluate the performance of the open dataset for sleep apnea detection. From the open dataset and the final test set, we randomly discard  four recording parts (a02,a03,c02,b01).

The Adversary trains a  neural network  to be able to differentiate the data from individuals, and attempts to find associations with the recordings in the open dataset. Meanwhile, a User utilizes the open dataset to train a model for the task of sleep apnea detection and generalizes on a test set. We explore three cases: Case (1): the open dataset includes  the raw data; Case (2): the open dataset includes AM stimuli;  Case (3): the open dataset includes  selectively inhibitory AM  stimuli based on the extension from Section \ref{sec:Extension}. For all models, we use a similar small convolutional-fully connected architecture. For more details please refer to Appendix A.

\begin{figure}
\begin{subfigure}{0.31\textwidth}
\includegraphics[width=\linewidth]{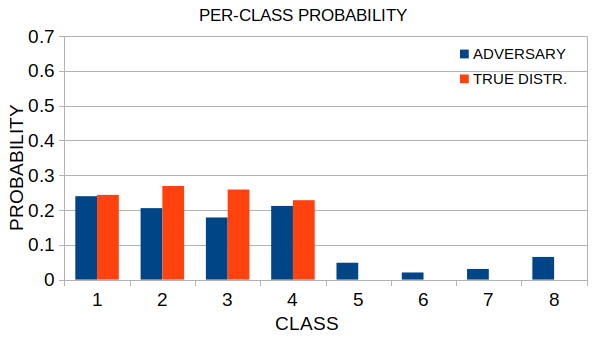}
\caption{Real Data} \label{fig:1b}
\end{subfigure}
\hspace*{\fill}
\begin{subfigure}{0.31\textwidth}
\includegraphics[width=\linewidth]{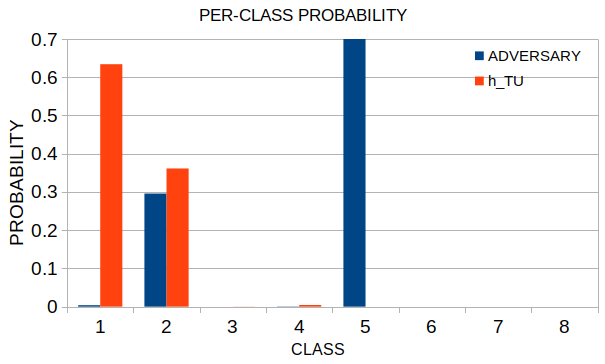}
\caption{AM stimuli} \label{fig:1b}
\end{subfigure}
\hspace*{\fill} 
\begin{subfigure}{0.31\textwidth}
\includegraphics[width=\linewidth]{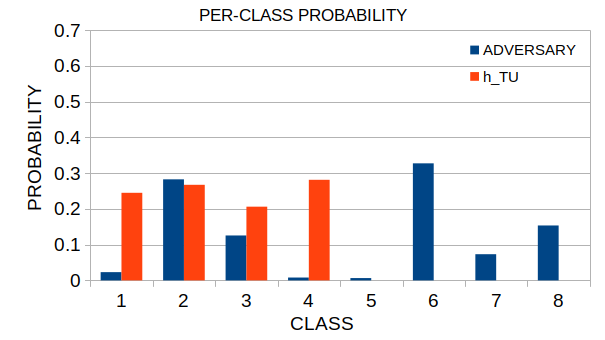}
\caption{AM inhibitory stimuli} \label{fig:1c}
\end{subfigure}

\begin{subfigure}{0.31\textwidth}
\includegraphics[width=\linewidth]{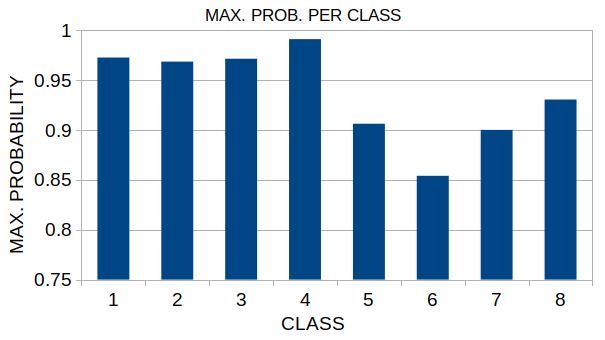}
\caption{Real Data} \label{fig:1b}
\end{subfigure}
\hspace*{\fill}
\begin{subfigure}{0.31\textwidth}
\includegraphics[width=\linewidth]{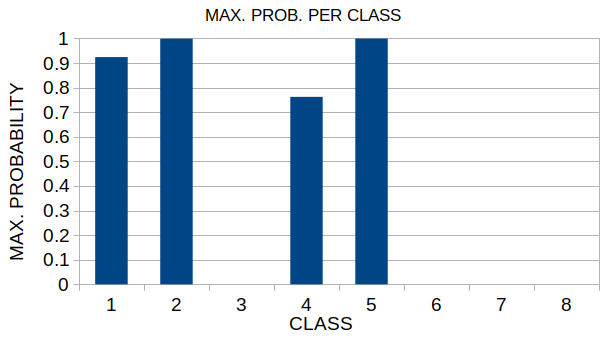}
\caption{AM stimuli} \label{fig:1b}
\end{subfigure}
\hspace*{\fill} 
\begin{subfigure}{0.31\textwidth}
\includegraphics[width=\linewidth]{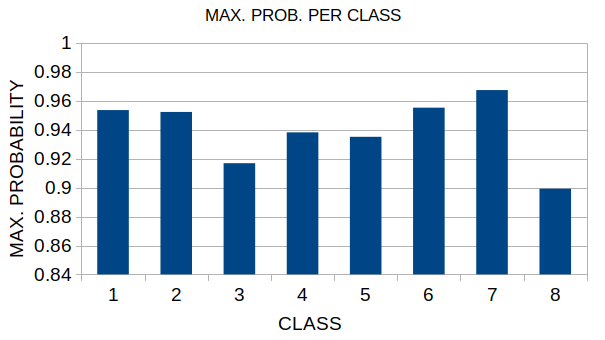}
\caption{AM inhibitory stimuli} \label{fig:1c}
\end{subfigure}
\caption{ (a-c): Per-class  histograms of the adversary  classification (blue) and the true labels (orange) for the three cases. (d-f): Maximum output probability per-class (averaged across the samples of the class). The four first classes correspond to the individuals of the open dataset. In all cases the depicted run corresponds to the run of median performance.} \label{fig:AttackRes}
\end{figure}

We need to evaluate two  properties in this experiment for the three cases. First, we need to evaluate the success of the Adversary. This is not trivial, and depends on what we mean by ``success''. For Case (1), we  calculate the performance of the adversary in classifying correctly individual minutes i.e., to which individual each minute belongs. We use the kappa statistic as the evaluation metric. For Cases (2) and (3),  we do not have direct access to the classes of the stimuli. Thus, we need a different  method. We approximate the stimuli classes with $h_{TU}$, i.e., a classifier  which belongs to the creator of the open dataset, that is trained to differentiate between the recordings in the open dataset. We  calculate kappa for Cases (2) and (3) with $h_{TU}$'s beliefs as true labels.  To gather more information, we  show in Figure \ref{fig:AttackRes} classification histograms of the adversary for the three cases we investigate, and the maximum output probability per class (average across all datapoints belonging to the specified class). These graphs can be used by the attacker as tools to extrapolate recording associations, so we show them for completion. Second, we need to evaluate the success of the User on the original task,  i.e., sleep apnea detection for each of the three cases. To do this we calculate kappa for the 3 cases on the test set.   All experiments are repeated ten times.

The  adversarial identification kappa is: 0.766, 0.251, and 0.004 for Cases (1),(2), and (3) respectively. As expected for the last two cases the adversary has  on average less success in recognizing correctly the individuals from which the datapoints indirectly occur than for Case (1). Figure \ref{fig:AttackRes} (a-c)  depict the per-class  percentage of classified datapoints of the adversary for the three cases. We also include the true (or approximated by $h_{TU}$) class distribution. The open dataset recording classes are the first four classes, i.e., classes 0-3.
For Case (1), the first four classes have much higher percentages. This is not true for Cases (2) and (3),  where the adversary is not able to successfully generalize to the appropriate classes. A difference between Cases (2) and (3) is that for Case (2) the percentages are much more concentrated   than for Case (3), which also holds for the approximated  ``true labels''. This is expected due to the inhibitory optimization with $h_{TU}$, through which we explicitly avoid the representations that $h_{TU}$ utilizes to distinguish between different  recordings. Figure \ref{fig:AttackRes}, (d-f) show the maximum output probabilities per-class. Again, for Cases (2) and (3), we do not observe  strong associations to the correct recordings. However, for Case (1)  the classes of the open dataset are more strongly represented than the other classes. For Case (2) some outputs are missing, because the adversarial predictions did not contain any data with maximum output of this class.
The kappa of the sleep apnea detection task on the test set is:  0.98, 0.903  and  0.968 for Cases (1),(2), and (3) respectively. Interestingly, Case (2), performed on-average worse than Case (3) despite the additional constraints imposed in the feature space for Case (3). 

From the above analysis we observe that both proposed procedures, and especially the Inhibitory AM (see Section \ref{sec:Extension}),  can successfully create an anonymized dataset, with relatively minor sacrifices in performance. 

\subsubsection{ EEG-Based Sleep Stage Classification: Comparisons With Related Works}
We repeat  the previous experiment on the larger Sleep EDF open dataset and compare with DP and non-DP variants of VAE and WGAN.  We use 21 recordings from the Sleep EDF Dataset (corresponding to the 21 first  patients).   Each one of these 21 recordings corresponds to a different patient. We use this number of patients because we want to achieve a good trade-off between data inclusion and adversarial success. For example, if we had included all 197 recordings, it would have been much harder for the adversary to perform the association attack. Empirically, including 21 recordings achieves this trade-off to a satisfactory degree.  
 The large EEG dataset allows us to hold -
out some patients and investigate a more realistic case than the previous OSA experiments. As such, we hold-out the first two patient recordings as test set for the task (i.e., sleep stage classification). 
We include all of the other 19 recordings in the Breach dataset, and the first 10 of the 19 recordings in the Open dataset.
To emulate the situation that the Breach and Open datasets include for each individual different recordings, we separate each recording into two halfs. The first half is included in the Open dataset if it belongs to one of the first 10 recordings.  The  second half is included in the Breach dataset.

In this experiment, we compare the AM approach with the WGAN and VAE. We include DP variants of these generative models, and experiment with different $\epsilon$ values. We additionally include the \textit{Censored Fair Universal Representations } (CFUR) \cite{kairouz2019censored}  generative method as a comparison baseline. CFUR comprises a set of methods in which a generative encoder encodes variants of the real data such that it competes against  a computational adversary which is a classifier. This adversary attempts to identify and classify successfully the sensitive information (e.g., in our case the patient), similar to our $h_{TU}$ classifier.  The training regime is adversarial, and the generator/encoder, attempts, given a distortion constraint $D_{ct}$, to make the adversary misclassify the encoded data. The overall procedure is very similar to the procedure followed by Feutry et al. \citeyear{feutry2018learning}, which we also discuss in  Related Works section. We  include this method as an additional baseline in our experiments since its defence mechanisms are by definition  better-suited against the proposed attack  than DP-based generative models. Therefore, such a method could give us better insights into what constitutes a proper defence given the specified scenario. Thus,  we investigate three more cases, in addition to the three cases from the previous section: Case (4) corresponds to WGAN experiments  Case (5) to VAE experiments, and Case (6) to CFUR experiments.
 
 As in the previous experiment, all synthesized data do not have labels for the $\mathbb X\rightarrow \mathbb U$ task. Therefore, we again retrain $h_{TU}$,  with the open dataset  to label the synthesized data and  to approximate $(p(u|x))_{u\in\mathbb U}$. To measure the performance of the adversary we compare the  agreement of the adversary  and $h_{TU}$'s labels for all the cases where synthetic data are synthesized by a generative model. We use for all  models (i.e., $h_T, h_{TU}, G_{AM}$, $G_{WGAN},D_{WGAN},E_{VAE}, D_{VAE}, E_{CFUR}$)  MLP architectures since more sophisticated architectures did not achieve substantially better performance. Furthermore,  for scenario-specific models, i.e.,  the User and the Adversary, we  mainly use MLPs. To evaluate the success of the adversary, and to understand the impact of the model architecture on this success, we use in addition to the MLP, a  convolutional-fully connected Adversary. We refer to the MLP Adversary as A-1 and the convolutional Adversary as A-2.  We tried to use similar architectures  for all models.  We describe the  architectures and hyper-parameter details in Appendix A.   
 All experiments are repeated 10 times, and the kappa statistic is used as performance metric. We generate 20000 synthetic datapoints for all models.
 
 Regarding the DP experiments, we use the  Renyi DP framework  \shortcite{mironov2017renyi} to calculate the privacy spent to train our models. Specifically, we use the $compute\_ rdp()$, and $get\_privacy\_spent()$ functions of  the tensorflow implementation. 
 Furthermore, we train our models with  the tensorflow implementation of the DP version of the Adam optimizer (which includes gradient clipping and addition of gaussian noise per-sample). We experiment with $\epsilon$ values of $\epsilon=100,10,3$ and $\delta=10^{-5}$. Originally we aimed for stronger DP guarantees in our experiments. However, we noticed a steep drop in task performance, especially in the case of the WGAN.  
 
 Regarding the CFUR experiments, we use Algorithm 1 from Kairouz et al. \citeyear{kairouz2019censored}, and  utilize the described   penalty method  to incorporate the distortion constraint into the learning algorithm.  We use three different distortion values for the distortion parameter $D_{ct}$, namely $D_{ct}=0.5,1,2$.
 
 Regarding the AM stimuli, for $T_Y$ the minimum sample in the training set has probabilistically  lower maximum probability than 0.5 in the output of $h_T$. Therefore,  we adopt as minimum $T_Y=0.5$, to properly perform AM. Regarding $T_U$, we use as $T_U=0.4$. We include more details about the other parameters of AM-Generation in Appendix E.

\begin{table}[h]
\footnotesize
\centering
\begin{tabular}{ |p{0.8cm}|p{1.6cm}|p{1.6cm}|p{1.6cm}|p{1.6cm}|p{1.6cm}|p{1.7cm}|p{1.6cm}|p{1.6cm}|}
\hline
\multicolumn{8}{|c|}{ EEG Sleep Stage Classification- Association Attack (Kappa). DP: ($\epsilon,\delta)=(10,10^{-5})$, Distortion:$D_{ct}$=1} \\
\hline
&Real Data& AM-I & WGAN & VAE& DPVAE&  DPWGAN &CFUR\\
\hline
Task&0.827$\pm$0.002&0.769$\pm$0.006&0.609$\pm$0.027&0.387$\pm$0.005&0.360$\pm$0.026&0.108$\pm$0.022&0.427$\pm$0.051\\
A-1 &0.414$\pm$0.004&0.020$\pm$0.003&0.270$\pm$0.020&0.263$\pm$0.024&0.165$\pm$0.009&0.048$\pm$0.021&0.231$\pm$0.043\\
A-2 &0.440$\pm$0.005&0.019$\pm$0.005&0.100$\pm$0.016&0.226$\pm$0.040&0.036$\pm$0.004&-0.02$\pm$0.001&0.109$\pm$0.042\\
\hline
\end{tabular}
 \caption{ Performance of the Sleep Stage Classification task  (Task) and the adversarial identification (A-1,2)  for the following data: Real Data,Inhibited AM Stimuli (AM-I), WGAN, and  VAE generated data, and DP-WGAN, DP-VAE and CFUR  generated data.}
 \label{table:EEG_1}
 \end{table}
 
 Tables \ref{table:EEG_1}, and  \ref{table:EEG_2} present  comparative results of  different variants of the generative models  on the task of sleep stage classification, and the identification success of A-1 and A-2. Table \ref{table:EEG_1} presents the results for the main chosen configurations of each model. As expected, the real data yield the highest performance, but also are the most susceptible to adversarial identification. We notice a clear advantage of the use of AM-I in comparison to the use of WGAN and VAE. Specifically for the WGAN case, even though the results for Task are relatively good, we notice that A-1   yields also high performance. For VAE the performance on the Task is relatively low and the performance of the Adversaries relatively high. As such the VAE does not seem to provide a very reliable solution. The introduction of DP increases as expected the protection against the adversarial identification, but at the cost of lower performance on the main task. Additionally, we observe that CFUR with distortion $D_{ct}=1$ yields on average  superior performance than the DP variants, and A-1, A-2 have generally inferior identification performances than the non-DP variants. Again we notice that AM-I seems to yield a superior trade-off.   In most cases, A-1 is higher than A-2, which is likely to happen  due to the architectural similarity between $h_{TU}$ and A-1.

  \begin{table}[h]
\footnotesize
\centering
\begin{tabular}{ |p{0.6cm}|p{1.65cm}|p{1.8cm}|p{1.65cm}|p{1.65cm}|p{1.65cm}|p{1.65cm}|p{1.65cm}|}
\hline
\multicolumn{8}{|c|}{Additional Results (Kappa) DP1: ($\epsilon,\delta)=(3,10^{-5})$, DP2: $(100,10^{-5})$, Distortion: $D_{ct1}=0.5$, $D_{ct2}=2$}  \\
\hline
& AM & DP1:WGAN & DP1:VAE& DP2:WGAN&  DP2:VAE& $D_{ct1}$:CFUR & $D_{ct2}$:CFUR\\
\hline
Task&0.704$\pm$0.011&0.043$\pm$0.038&0.272$\pm$0.026&0.233$\pm$0.017&0.356$\pm$0.020&0.652$\pm$0.017&0.196$\pm$0.053\\
A-1 &0.091$\pm$0.002&0.065$\pm$0.030&0.184$\pm$0.007&0.074$\pm$0.024&0.161$\pm$0.010&0.296$\pm$0.033&0.022$\pm$0.036\\
A-2 &0.016$\pm$0.001&0.036$\pm$0.016&0.036$\pm$0.004&0.000$\pm$0.012&0.019$\pm$0.005&0.129$\pm$0.050&0.001$\pm$0.027\\
\hline
\end{tabular}
 \caption{Performance of AM (no Inhibition), WGAN,  VAE, and CFUR with stronger and weaker privacy guarantees.}
 \label{table:EEG_2}
 \end{table}
 
 Table \ref{table:EEG_2} provides results from additional configurations we explored.  A-1's performance on the VAE data does not seem to drop when we strengthen the privacy guarantees from $\epsilon=100$ to $\epsilon=10$ or 3.  The AM column investigates how the use of  AM stimuli without inhibition performs. Again, we notice  good behaviour. However A-1 yields superior performance than in the AM-I case. Interestingly, also AM-I yields higher Task performance.   Furthermore, we observe that CFUR with $D_{ct}=0.5$ yields superior performance than $D_{ct}=1,2$ at the expense of higher identification performance from A-1 and A-2. The opposite holds for $D_{ct}=2$. This behaviour is expected as, increasing $D_{ct}$ practically increases the distortion range of $E_{CFUR}$, i.e., its capability to ``fool'' the computational adversary of CFUR.

 Overall, based on the above results, we identify that  AM and especially AM-I provide good trade-offs between anonymization and performance, which seem to be superior to the other investigated approaches, for this dataset. Regarding the DP cases, we observe based on our experiments and related literature \cite{hitaj2017deep} that the DP framework potentially exhibits partial susceptibility to this type of associative identification attacks.    We hypothesize that a reason for this phenomenon  is that the DP framework is potentially less  suited for such a population-based attack than direct inhibition on the feature space. For a DP framework to provide guarantees on a population level (i.e., in our case multiple datapoints comprising a whole recording), we must reach much higher privacy levels due to the group privacy property. However, this  option is not  viable  since,  in our experiments,  already with the current DP levels, the performance of the DP based models suffers in comparison to non-DP variants. Regarding CFUR, we hypothesize that a potential explanation could relate with the higher amount of available movement in the input space with AM/AM-I. The proposed techniques are not bounded to specific datapoints. This (especially for the case of AM-I) could potentially enable task related features to be learned indirectly and independently of sensitive identification features as the stimuli can navigate away of sensitive regions of the input space.

  \subsection{Student Resilience Against Membership Inference Attacks}
An additional side effect of the proposed approach  is the resilience $h_S$ obtains against certain information leakage attacks. This phenomenon occurs since $h_S$ is trained with a substitute training dataset, whose marginal distribution is different from that of the  original data. In this section, we experimentally evaluate the resilience of $h_S$ against  membership inference (MI) attacks.

 In MI attacks, the adversary tries to determine whether a given datapoint belongs to the training dataset, based on  the output  of $h_S$. For a classifier $h_T$ trained on the real training data, we expect a statistically different behavior on its output when presented with data that were used to train it than when presented with data that were not used to train it. This difference can be exploited by an adversary to learn whether a datapoint belongs to the training set or not. However,  since we use a very different dataset  from the real training  dataset which also follows different distribution, we hypothesize that it is possible to negate the effect of the MI attack for $h_S$. We follow in this experiment the approach from Shokri et al. \citeyear{shokri2017membership}, but  we simplify it. Instead of using shadow models trained on  synthetic data to imitate the performance of the target classifier ($h_S$), we assume that we have auxiliary information in the form of a-priori knowledge of whether data belongs to the training or test set of the target model. Thus, we train a new classifier $h_{mi}$ on the ouput of $h_S$ to recognize whether a datapoint belongs to the training set of $h_S$ or not. We evaluate the success of the attack with the use of kappa statistic since this is a two class problem (belong to train or to the test set), and we use the Apnea-ECG and A3 datasets for evaluation.  Additionally, we apply the attack  to $h_T$,  i.e., the Teacher, which is a similar model to the Student classifier $h_S$ that has been trained with real data instead of stimuli. Thus, we compare how successful the attack is when  applied to (1) a classifier which has been trained directly with the real data and (2) to $h_S$ i.e., a classifier trained with the AM stimuli in place of the real data.  We use the   $ID$ architecture from the  experiments of Section 5, as $h_S$ and  $h_T$.  We use for $h_{mi}$ a small 4-layer fully connected network with 30-50 neurons per layer and  relu activations. We train $h_S$ and $h_T$ for Apnea-ECG and A3 for 100 and 10 epochs respectively. For both datasets we do not use the whole dataset but a randomly drawn sample of size 1000 for Apnea-ECG (500 training and 500  test) and 15000 for A3 (7500  training and 7500  test). In both cases we use 66.6\% of the dataset to train $h_{mi}$ and 33.3\% to evaluate. The experiment is repeated five times. 
 
The Apnea-ECG  results, in terms of kappa   are: $h_T$: 0.120, $h_S$: 0.044.  For the A3 study: $h_T$: 0.231, $h_S$: 0.042. These results correspond to  the  success of $h_{mi}$ in identifying whether  previously unseen data  were or were not used to  train the model which is under attack ($h_T$ or $h_S$ depending on the experiment). Though $h_S$ does not directly use the training data during its training, the training-test separation is the same for $h_S$ and $h_T$. From these results we observe that even though   the attack is  for   datasets relatively unsuccessful both for $h_T$ and $h_S$,  it is much more successful for $h_T$ than for $h_S$, as expected.  Despite this, please note that the proposed approach falls into the utility-privacy trade-off commonly identified when using such anonymity methodologies, since in practice we sacrifice classification performance to achieve anonymity for a given task ($\mathbb X\rightarrow \mathbb U$) and to hinder possible associations with the real data via producing visually unrealistic datapoints. It is important to note that membership inference attacks are not directly designed to yield good results against generative approaches like the proposed. As such, our goal in this section is to provide an  experimental scenario that investigates the resilience of a student when trained on stimuli compared to real data, in the context of a common real-world attack. 

\subsection{ Class Recognition of Stimuli }

An interesting  observation regarding the proposed approach is that there seems to be disagreement across different algorithms trained on the real data, for the  classes of the stimuli. This occurs even  for the $\mathbb X\rightarrow \mathbb Y$ task, for which the stimuli maximize $h_T$'s output.   We experimented with  a linear SVM, a DNN (relu), a Random Forest (RF) ensemble with 50 trees, a Convolutional architecture identical to $h_T$  and three humans.  We trained  the four algorithms on the real data and  evaluated them on a batch  of 10000 from the MNIST stimuli dataset. For the humans, we randomly extracted a smaller batch of 100 stimuli and they performed digit classification on the batch. We used another 100 MNIST real test datapoints as baseline. The results are shown in Table \ref{table_Rec}. The examined methods and humans are not able to sufficiently identify the  AM stimuli. As before, we attribute this to the inductive bias mismatch between the methods or humans  and $h_T$. This is also supported by the much better recognition for the  convolutional architecture identical to  $h_T$. Notice also the strong correlation with Section \ref{OtherCl}. The models that had more difficulty to learn the patterns and generalize successfully to the real data are the ones performing the worst also in this experiment, as expected. 

  \begin{table}[h]
\centering
\begin{tabular}{ |p{1.8cm}|p{1.5cm}|p{1.5cm}|p{1.5cm}||p{1.5cm}||p{1.5cm}|}
\hline
\multicolumn{6}{|c|}{ Stimuli Class Recognition  MNIST ($h_T$ trained for 2344 batches, Acc\%)} \\
\hline
Method& SVM& RF& DNN&Humans&Conv $ID$  \\
\hline
Baseline&93.91&96.69&97.69&97.00&99.20\\
AM Batch&26.78&28.85&38.84&17.30&79.39\\
\hline
\end{tabular}
\caption{Class recognition of the AM stimuli from different algorithms trained on the real data and three humans. It is measured as accuracy over the stimuli batch. }
 \label{table_Rec}
\end{table}

\section{Discussion and Limitations}
\label{sec:Discussion1}

In this section we analyze  important observations from our experiments, and discuss potential weaknesses of the proposed approach.

  \subsection{Training Times}

 An interesting point of discussion is the required training time of the proposed approach. AM is by definition efficient, since assuming that we can easily reach the specified threshold, the AM generator only needs several steps to converge and satisfy the predefined condition. 

 \begin{wrapfigure}{r}{0.5\textwidth}
  \begin{center}
    \includegraphics[width=0.48\textwidth,height=5cm]{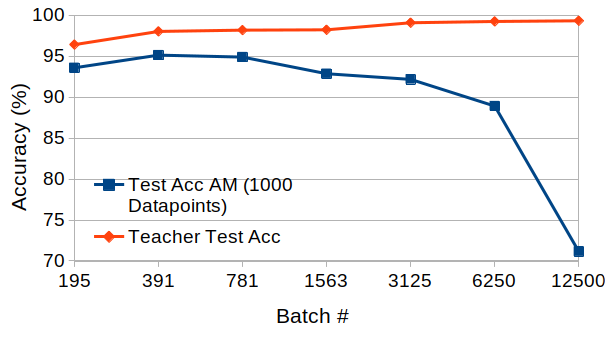}
  \end{center}
  \caption{Accuracy of $h_T$ (orange) and ID $h_S$ (blue) for 1000 stimuli, for different number of training iterations of  $h_T$.}
\label{fig:TestAM}
\end{wrapfigure}

\noindent The total time however  accumulates linearly with the number of stimuli. 

 In this work, we did not focus on training time, and we experimented mainly with larger numbers of stimuli. However, we can still obtain satisfactory performance even with a relatively small number of  stimuli, e.g., with 3000 stimuli we can  achieve an accuracy of $92$\% on MNIST. We needed close to 360 seconds to generate 3000 stimuli on a Nvidia-RTX 2080 Graphics Card.

In our experiments, given the chosen configurations from our pre-study, VAE required 1076 seconds and WGAN required 2628 seconds to train on the MNIST dataset. \footnote{From our pre-study, we chose near 55 training epochs for WGAN, and 100 epochs for VAE. A batch size of 128 was chosen in both cases.} As such, we observe that it can be more efficient to generate a relatively small number of stimuli than training a new generative model if training time is a priority. On the other hand, if we want to generate a relatively large dataset  it could be more efficient in terms of training times to train once a generative model, instead of producing multiple independent stimuli.   Based on this discussion the overall training time  depends on how many stimuli we need, which   in turn depends on our performance requirements. 

Additionally, the generalization capability of $G_{AM}$ can be leveraged    to generate multiple stimuli after the AM generator has converged. However, the stimuli from each run are very similar. Thus,  additional diversity regularization is required. We leave this option for future work.

Another important point regarding training times is related to data dimensionality. We  empirically observe that as data dimensionality increases the time needed to generate one stimuli becomes relatively shorter in comparison to the time needed to train a new generative model. For example, generating a stimuli dataset of  EEG sleep stage data, which has lower dimensionality than the  MNIST data, was more inefficient than for the MNIST case, when compared to training a generative model. We attribute this characteristic to the relative simplicity of the AM task. As we scale-up data dimensionality, the AM task  can potentially become relatively easier to accomplish than in lower dimensionalities when compared for example to a more complex adversarial objective. However, more work is needed to verify and generalize this observation. 

\subsection{Infrequent Patient Data in the Real Dataset }

An important consideration of the proposed approach regards the relative frequency of the data of each patient in the real dataset. For example, if a patient has contributed less data relative to the other patients that are included in the dataset, a question arises about whether $G_{AM}$ can produce appropriate inhibitory stimuli that can inhibit patient-specific features for the particular patient. This consideration arises since $h_{TU}$ has less data to work with for the class of the particular patient. As such it has potentially learned weaker representations relevant to this class, which could hinder $G_{AM}$'s learning.

However, as $G_{AM}$ inhibits the output of $h_{TU}$ towards the empirical marginal, a counter-argument can be made since the relative frequency of each patient's data is empirically mapped onto $\hat{p(u)}$. As such, $G_{AM}$ navigates  during the stimuli generation  by definition  towards values of lower probability for the infrequent classes, thus making it easier for $G_{AM}$ to generate inhibited stimuli despite the weaker class representation imposed on $h_{TU}$ for the particular patient.
Despite this discussion, this work has served as a first step proof-of-concept for the proposed approach, and as such we generally, used well-balanced datasets  in terms of patient representation. Based on this we have not accumulated sufficient empirical evidence to support any of the abovementioned hypotheses. We leave  further experimentation with unbalanced datasets in terms of patient representation for future work.

\subsection{Performance Per Stimulus}

We observe that the performance per stimulus is not the same among   $h_T$  trained for different numbers of epochs. Figure \ref{fig:TestAM} shows for MNIST the performance of   $h_T$  trained  for different  numbers of epochs on the orange line. The blue line depicts how the $h_S$ with identical architecture performs when trained with 1000 data stimuli generated from the respective  $h_T$. We observe that when  $h_T$  is trained for a larger number of batches, the performance of $h_S$ for 1000 stimuli significantly drops.  The best performance of $h_S$ in this experiment is obtained only for 390-790 batches of training. However, for 390-790 batches  the performance of  $h_T$  is generally low, and so the maximum reachable performance  for $h_S$ is lower than for the $h_T$ that are trained for more iterations. For this reason, we choose the  $h_T$  trained for 3125 batches  for our MNIST experiments. This choice yields a good $h_T$ performance and not the worst  AM dataset performance for 1000 stimuli. We use similar criteria for the training of $h_T$ for the apnea datasets.

  \subsection{Connection to Adversarial Objectives and Domain Adaptation}
  
 An interesting connection can be made between the proposed approach and adversarial objectives. We use $h_T$ excitation to produce stimuli with which  task $\mathbb{X}\rightarrow \mathbb{Y}$ can be learned from a new student. At the same time we  de-identify the stimuli with the use of $h_{TU}$. This objective is not adversarial, in the sense that the two tasks are different but not necessarily opposing.
  However, depending   on the tasks  $\mathbb{X}\rightarrow \mathbb{Y}$ and $\mathbb{X}\rightarrow \mathbb{U}$, we hypothesize that hidden “adversarial” aspects between the two tasks could potentially emerge, depending on the context of these tasks.  
  
  For example, given the scenario from Section 2.2.4, if $\mathbb{X}\rightarrow \mathbb{U}$ corresponds to person identification, and $\mathbb{X}\rightarrow \mathbb{Y}$  to identification of people with long hair, let us assume that  Tom is the only person in our data who has contributed images with long hair. Then generating stimuli from $h_T$ for $\mathbb{X}\rightarrow \mathbb{Y}$ on the class of identified long hair can potentially be problematic for the inhibition of features of Tom, since $h_T$ could have associated features of Tom to positive long hair identification. Then, inhibition of $h_{TU}$ ``away'' from Tom, could be adversarial towards stimulation of $h_T$ towards ``long hair''.  Such cases of emerging adversity in the objective can be problematic  for the proposed approach. The investigation of such correlations between two tasks is an interesting topic for future work.
  
  Furthermore, an additional connection can be made between the proposed approach and domain adaptation, in the sense that we introduce a loss such that we prevent the generative model ($G_{AM}$) to learn domain- specific information for $\mathbb{X}\rightarrow \mathbb{U}$.  
 
 \subsection{General Risks}
 
 It is important to mention that even though the  stimuli inhibition is supported by theoretical insights, the proposed approach does not constitute a general privacy solution that can ameliorate all types of privacy leaks. For example, as we focus on the inhibition of $\mathbb X \rightarrow \mathbb U$, defense against other potential orthogonal identification tasks can only occur from the inherent  misidentification potential of the stimuli. As such, we agree with the conclusions from Bellovin et al. \citeyear{bellovin2019privacy} in the sense that  the proposed synthetic data generative approach, as the majority of  solutions, does not offer a ``silver-bullet''. Instead,  based on our empirical results, it can potentially be used for specific anonymization purposes.

Since this is based on empirical evidence we can not exclude that in the future new attacks might be developed for which our solution is vulnerable.

\section{Related Work}
\label{sec:rw}

We first discuss related works  which regard transferring knowledge between different classifiers. Then we  turn our focus on  anonymization for generative models. Finally, we analyze additional works that are useful and provide insights towards our approach. 

\subsection{Knowledge Transfer}

Recently, many new techniques for transferring knowledge have   been  proposed,
especially with the goal of reducing the size of a DNN to decrease the execution
time and reduce memory consumption. Existing model compression techniques,
e.g., via pruning or parameter sharing \shortcite{cheng2017survey,hassibi1993second,han2015learning,srinivas2015data} can be considered as a form of
knowledge transfer from a trained teacher to a student. Other types of methods perform 
  knowledge transfer  between different tasks and domains \shortcite{yim2017gift,pan2010survey}. Furthermore, techniques exist that transfer knowledge  from  smaller DNNs to  equal sized or larger DNNs to make them learn faster 
\shortcite{chen2015net2net} or to perform better than the original network \shortcite{furlanello2018born}.
In the knowledge distillation method \shortcite{hinton2015distilling}, the student network is trained
to match  the classes of the original data  together with a modified version of the softmax output probabilities  from the trained teacher network.  This allows to control  the steepness  of the output  class probability distribution. A similar approach is mimic learning \shortcite{shafee2020mimic}, where a Teacher is trained on a dataset comprising sensitive data. Then the Teacher is released and labels an unlabelled real public dataset, and a Student  is trained on the labelled public dataset. Romero et al. \citeyear{romero2014fitnets} introduce fitnets, an extension of 
knowledge distillation  to train thinner deeper networks (student) from
wider shallower ones (teacher). Bucilua et al. \citeyear{bucilua2006model} investigate the compression of large ensembles
(like RF, bagged decision trees, etc.) via the use of a very small artificial neural
network (ANN). As a universal approximator, the ANN is able to generalize
to mimic the learned function of the ensemble. To train
the ANN they create a larger synthetic dataset based on the real dataset that
is labeled by the ensemble. Luo et al. \citeyear{luo2016face} use knowledge distillation on a selection of
informative neurons of top hidden layers to train the student network. The
selection is done by minimizing an energy function that penalizes high correlation
and low discriminativeness.  Papernot et al. \citeyear{papernot2016semi} use a similar student-teacher(s)  scenario in order to train a student in a differentially private manner.  The teacher ensemble is not released to the user, and also training of the student is done with real data (that are not in the private dataset) and a GAN used in a semi-supervised manner based on the method proposed by Gulrajani et al.  \citeyear{gulrajani2017improved}.

\subsection{Generative Models and Anonymity}
 Many anonymization approaches for generative models exist which incorporate the DP framework into the GAN framework \shortcite{jordon2018pate,xie2018differentially,xu2019ganobfuscator}. Contrary to these approaches, we focus on ``hiding'' certain  data features. We  achieve this by learning to avoid representations that leak these features, based on the output of an approximator of the true conditional. Thus, the proposed approach is data dependent and  not a universal method incorporated in a mechanism.  Other approaches manipulate the GAN framework in order to generate anonymized  data (mainly images) for medical or general purposes  \shortcite{shin2018medical,hukkelaas2019deepprivacy}. Several approaches \shortcite{bae2019anomigan,feutry2018learning,kairouz2019censored} utilize  data-centric anonymization strategies similar to ours in order to hide sensitive information from potential adversaries. However,   these aforementioned works modify or encode existing datapoints with the use of adversarial frameworks. In contrast, we generate stimuli without directly accessing the real data. An interesting observation is that the vast majority of  recent works utilizes modifications of the GAN framework to achieve their respective goals. The GAN framework is a natural choice in cases where data realism is important. As we do not have such a requirement we are interested in exploring an alternative approach.  However, it is also important to mention that generating unrealistic data in this context is not the final objective. Instead, we hypothesize that  by not being bound to generating realistic data, we  have more freedom to explore different, potentially promising  solutions.

\subsection{AM and Other Works}
\par To better  understand the internal representations of a DNN,  the AM technique is introduced by Erhan et al. \citeyear{erhan2009visualizing} for qualitative
evaluation of higher-level  internal representations of two
unsupervised deep architectures. Nguyen et al. \citeyear{nguyen2016synthesizing} use  a deep
generative network   to synthesize images that maximize the output of a neuron of
a certain layer of the network. We base our approach on some of these insights.

For the purpose of clarity, a distinction  should be made between our approach and approaches which include the use of trained  convolutional filters.   Convolutional filters are good feature extractors \shortcite{kiela2014learning,garcia2018behavior}  and these features can be used to train  models \shortcite{zeiler2014visualizing}, or be transferred from the feature space of a given domain to another domain \shortcite{xie2016transfer}.  However, we focus on training (from scratch) a new student classifier so that it can generalize well to a sub-region of the feature space defined from  the AM (maximization or minimization). To do this, we take advantage of the same inductive bias which occurs by the use of a  similar student architecture to that of the teacher. We do not use teacher layers as feature extractors. We use synthetic data sampled from the marginal distribution after AM is performed and not real data from another domain to train the student.

\section{Conclusion and Future Work}
\label{sec:concl}

The primary motivation for the proposed approach is to   address  the problem of  limited   training data availability   due to anonymity and sharing regulations. Our aim is to enable  users to  successfully  train and customize models while minimizing the risk of identification  for individuals who have contributed  in the  formation of the original dataset.  As such, arbitrary users can benefit from anonymized medical data sets to train or develop their own classification models. To achieve this we  utilize AM in a generative manner, and   create a multi-faceted dataset of stimuli that captures implicitly the class separation of the real data.  

In this work, we emphasize application in a medical setting, and  apply the  proposed approach on the  problems of sleep apnea detection and sleep stage classification based on EEG.  We utilize data from real-world clinical studies, and   showcase its viability for these tasks. Furthermore, we evaluate  on the task of digit recognition, and verify that the proposed approach is generalizable across different tasks and domains. Training with synthetic stimuli  can yield promising results that are comparable or superior to  well-established generative methods which can successfully produce realistic data. In this paper, we mainly evaluate  on smaller classifiers potentially for use in a resource constrained environment. We experimentally show that  we can utilize synthetic  stimuli in place of real data to  anonymize individuals that have  contributed to the real dataset with their data.  Furthermore, we  utilize the proposed approach to produce a classifier that is  more resilient against specific information leakage attacks, namely MI attacks.

In our ongoing and future work we address the customization of a student classifier $h_S$ towards the personal and unlabeled data of the end-user. In other words, we aim to use NE for domain adaptation with  only $h_T$ and the unlabeled data of the end-user as input, and a personalized $h_S$ as output.

\acks{ This work was performed as part of the CESAR project (nr.
250239/O70) funded by The Research Council of Norway. During this work Gunn Marit Traaen was also affiliated with the
Institute of Clinical Medicine, Faculty of Medicine, University of Oslo,
Oslo, Norway. We want to thank the anonymous reviewers for their
valuable feedback that helped us to substantially improve this manuscript.
}

\appendix
\section{ Architectures and Hyperparameters for the Models Used}

In Appendix A, we show the architectures and hyperparameter values for the different models used in the experiments. We present the architectures of (1) the trained networks containing  knowledge for the data distribution (conditional or joint, i.e., teachers T), (2) of the students that learn implicitly from the teachers, and (3) of other networks that are used as intermediates (namely the AM generator and the WGAN discriminator). Tables 3-8 showcase these results.

For Apnea-ECG, we use a batch size of 100. For A3 and MNIST, we use batch size of 128. All of random noise input are sampled from a standard normal distribution. The learning rate used for the Apnea-ECG and A3 experiments is 0.0001, and 0.001 for the MNIST experiments. We omit biases in all tables for simplicity. In all cases they correspond to the size of the output for the layer (number of channels on Convolutional or Transpose convolutional Layers).

\subsection{ Teachers T}
We present the architectures of the models used as T for the different experiments in Tables \ref{table:ArchsT1},\ref{table:ArchsT2},\ref{table:ArchsT3}. 

\begin{table}[h!]
\centering
\small
\begin{tabular}{ | p{3.75cm}| p{3.75cm}|p{3.75cm}|}
\hline
\multicolumn{3}{|c|}{Architectures used} \\
\hline
 VAE Decoder& WGAN Generator& Classifier $h_T$($ID$ Arch.)\\
\hline
fc,D,lrelu: 12$\times$60&fc,D,lrelu: 31$\times$60& fc,D,relu: 60$\times$360  \\
fc,D,lrelu: 60$\times$180&fc,D,lrelu:60$\times$180& fc,D,relu: 360$\times$180 \\
fc,lrelu: 180$\times$360&fc,D,tanh:180$\times$360& fc,relu: 180$\times$64 \\
fc,$\lambda$*tanh: 360$\times$60& fc,$\lambda$*tanh:360$\times$60&fc,softmax: 64$\times$2\\
\hline
\end{tabular}
\caption{ Teacher Architectures used for the Apnea-ECG experiments (fc: Fully connected, input$\times$output. The activation function used is shown next to the layer type separated with comma. If used, we show dropout as D).}
 \label{table:ArchsT1}
\end{table}

\begin{table}[h!]
\footnotesize
\centering
\begin{tabular}{ | p{4.2cm}| p{4.2cm}| p{4cm}|}
\hline
\multicolumn{3}{|c|}{Architectures used} \\
\hline
 VAE Decoder&WGAN Generator& Classifier $h_T$ ($ID$ Arch.)\\
\hline
fc,relu: (10+10)$\times$3136&fc,relu: (128+10)$\times$1024& Conv,relu,MP: 1$\times$32$\times$5$\times$5 \\
ConvTr,relu: 64$\times$3$\times$3$\times$64&fc,relu: 1024$\times$2048&Conv,relu,MP: 32$\times$28$\times$5$\times$5\\
ConvTr,relu: 64$\times$3$\times$3$\times$32&ConvTr,relu: 128$\times$3$\times$3$\times$32&fc,relu,D: (28$\times$7$\times$7)$\times$1024 \\
ConvTr,sigmoid: 32$\times$3$\times$3$\times$1&ConvTr,relu: 32$\times$4$\times$4$\times$28&fc,softmax: 1024$\times$10\\
-&ConvTr,sigmoid: 28$\times$4$\times$4$\times$1& -\\
\hline
\end{tabular}
\caption{ Teacher Architectures used for the MNIST experiments (Conv/ConvTransp: input channels$\times$output channels$\times$filter, MP: Max Pooling, fc: Fully connected, input$\times$output).}
 \label{table:ArchsT2}
\end{table}

\begin{table}[h!]
\footnotesize
\centering
\begin{tabular}{| p{4cm} | p{4cm}| p{4cm}|}
\hline
\multicolumn{3}{|c|}{Architectures used} \\
\hline
 VAE Decoder&WGAN Generator& Classifier $h_T$ ($ID$ Arch.)\\
\hline
fc,D,lrelu: (20+2)$\times$120&fc,D,relu: (60+1)$\times$120& Conv,relu,MP: 1$\times$4$\times$1$\times$16 \\
fc,D,lrelu: 120$\times$15*64&fc,D,relu: 120$\times$1500& Conv,relu,MP: 16$\times$4$\times$1$\times$32\\
ConvTr,lrelu: 64$\times$4$\times$1$\times$64&ConvTr,relu: 100$\times$4$\times$1$\times$16& Conv,relu,MP: 32$\times$4$\times$1$\times$64 \\
ConvTr,lrelu: 64$\times$4$\times$1$\times$16&ConvTr,tanh: 16$\times$4$\times$1$\times$32&fc,D,relu: (64$\times$8)$\times$64\\
ConvTr,lrelu: 32$\times$4$\times$1$\times$1&ConvTr,tanh: 32$\times$4$\times$1$\times$64& fc,D,relu: 64$\times$32\\
-&ConvTr,tanh: 64$\times$2$\times$1$\times$1&fc,softmax:32$\times$2\\
\hline
\end{tabular}
\caption{ Teacher Architectures used for the A3 experiments (Conv/ConvTransp: input channels$\times$output channels$\times$filter, MP: Max Pooling, fc: Fully connected, input$\times$output).}
 \label{table:ArchsT3}
\end{table}

\subsection{ Students}
We present the architectures of the models used as students for the different experiments in Tables \ref{table:ArchsS1},\ref{table:ArchsS2},\ref{table:ArchsS3}. 

\begin{table}[h!]
\centering
\small
\begin{tabular}{ | p{2.4cm}| p{2cm}| p{2cm}| p{2cm}| p{1.5cm}| }
\hline
\multicolumn{5}{|c|}{Student Architectures used Apnea-ECG} \\
\hline
Classifier $h_S$ & $S$ Arch.&$VS$ Arch.&$L$ Arch.&$LL$ Arch.\\
\hline
 fc,D,relu:&60$\times$180&60$\times$90& 60$\times$360&60$\times$360  \\
fc,D,relu:&180$\times$90&90$\times$45& 360$\times$180& -    \\
fc,relu: &90$\times$32&45$\times$16& - & - \\
fc,softmax:&32$\times$2&16$\times$2&180$\times$2& 360$\times$2 \\
\hline
\end{tabular}
\caption{ Student Architectures used for the Apnea-ECG experiments (fc: Fully connected, input$\times$output. The activation function used is shown next to the layer type separated with comma. If used, we show dropout as D: Dropout).}
 \label{table:ArchsS1}
\end{table}

\begin{table}[h!]
\centering
\small
\begin{tabular}{ | p{2.4cm}| p{2.5cm}|p{2.2cm}|p{2.2cm}|p{2.2cm}|}
\hline
\multicolumn{5}{|c|}{Student Architectures used MNIST} \\
\hline
Classifier $h_S$ & $S$ Arch.&$VS$ Arch.&$L$ Arch.&$LL$ Arch.\\
\hline
 Conv,relu,MP:& 1$\times$16$\times$5$\times$5 & 1$\times$16$\times$5$\times$5& 1$\times$32$\times$5$\times$5&1$\times$32$\times$5$\times$5\\
Conv,relu,MP:& 32$\times$14$\times$5$\times$5 & 16$\times$12$\times$5$\times$5& 32$\times$28$\times$5$\times$5&-\\
fc,relu,D: &(14$\times$7$\times$7)$\times$512 &(12$\times$7$\times$7)$\times$32&-&- \\
fc,softmax: &512$\times$10 &32$\times$10&(28$\times$7$\times$7)$\times$10&(32$\times$7$\times$7)$\times$10\\
\hline
\end{tabular}
\caption{ Student Architectures used for the MNIST experiments (Conv/ConvTransp: input channels$\times$output channels$\times$filter, MP: Max Pooling, fc: Fully connected, input$\times$output).}
 \label{table:ArchsS2}
\end{table}

\begin{table}[h!]
\small
\centering
\begin{tabular}{ | p{2.3cm}| p{2.3cm}|p{2cm}|p{2cm}|p{2cm}|}
\hline
\multicolumn{5}{|c|}{Student Architectures used} \\
\hline
 Classifier $h_S$ &$S$ Arch.&$VS$ Arch.&$L$ Arch.&$LL$ Arch.\\
\hline
Conv,relu,MP:& 1$\times$4$\times$1$\times$8&1$\times$4$\times$1$\times$4& 1$\times$4$\times$1$\times$16& 1$\times$4$\times$1$\times$16 \\
Conv,relu,MP:& 8$\times$4$\times$1$\times$16&4$\times$4$\times$1$\times$8 &16$\times$4$\times$1$\times$32&16$\times$4$\times$1$\times$32 \\
Conv,relu,MP:& 16$\times$4$\times$1$\times$32 &8$\times$4$\times$1$\times$16&32$\times$4$\times$1$\times$64&- \\
fc,D,relu: &(32$\times$8)$\times$32&(16$\times$8)$\times$16&-&- \\
(MP),fc,D,relu: &32$\times$16&16$\times$8&(64$\times$8)$\times$32&(32$\times$8)$\times$32 \\
 fc,softmax: & 16$\times$2& 8$\times$2&32$\times$2& 32$\times$2\\
\hline
\end{tabular}
\caption{ Student Architectures used for the A3 experiments (Conv/ConvTransp: input channels$\times$output channels$\times$filter, MP: Max Pooling, fc: Fully connected, input$\times$output).}
 \label{table:ArchsS3}
\end{table}

\subsection{ Other Networks}
In this section we describe the additional intermediate architectures used in our experiments (i.e., WGAN Discriminators and the AM Generators).

\subsubsection{ AM Generator}
We experimented with a variety of configurations as AM generator for the different experiments:
\begin{itemize}
\item \textbf{Apnea-ECG:} We use a 7-layer fully connected (512, 256, 256, 180, 60) MLP with relu activations and dropout in the first two layers. The noise input vector has size 512.

\item \textbf{MNIST:} We use a convolutional-fully connected deconvolutional architecture, with 2 convolutional layers ($1\times$32$\times$5$\times$5, 32$\times$16$\times 3\times 3$), 1 fully connected ($7 \times 7 \times16\times 588 $), and 3 deconvolutional layers ($12\times 32\times 2\times 2$,$32\times 32\times 4\times 4$, $32\times 1 \times 4\times 4$). All activations are relu except for the last layer which is use sigmoid. the input noise vector has shape (28,28,1).

\item \textbf{A3}: We use a fully connected-deconvolutional architecture with a 512 input noise vector,   with 2 fully connected  (MLP) layers (512,1000)  and 5 deconvolutional layers (200$\times 128 \times 3 \times 1$, $128\times 128 \times 4\times 1$, $128\times 64 \times 4 \times 1$, $64\times 16 \times 5 \times 1$, $16\times 1\times 2 \times 1$).
\end{itemize}

\subsubsection{ WGAN Discriminator}
We experimented with a variety of configurations as AM generator for the different experiments:
\begin{itemize}
    \item \textbf{Apnea-ECG:} We use a fully connected MLP with 4 fully connected layers, and an input of 61 (a 60 dimensional feature vector +1 a one dimensional  condition) (120, 180,30, 1), with relu activations and dropout in the first 2 layers.
    
    \item \textbf{MNIST:} We use a convolutional network with an input of (28,28,1), 3 convolutional layers ($1\times 128 \times 5\times 5$, $128 \times 64 \times 5\times 5$, $64 \times 32 \times 5 \times 5$) and 2 fully connected layers (128, 1).   We use in all cases relu activations, and dropout in the second to last  fully connected layer. We pass the condition in the first fully connected layer.

    \item \textbf{A3:} We use a convolutional network with an input vector of 60, 3 convolutional layers ($1\times 64 \times 4\times 1$, $64 \times 32 \times 4\times 1$, $32 \times 16 \times 4\times 1$), and 3 fully connected layers (120, 30, 1). We use in all cases relu activations, and dropout in the 2 fully connected layers. We pass the condition in the first fully connected layer.
\end{itemize}
\subsubsection{VAE Encoder}

In all experiments, the VAE encoders have identical inverted architectures than the VAE decoder of the respective experiment. For example for the MNIST experiments  the encoder includes: Convolutional layer, Convolutional layer  Convolutional layer, fully connected layer. Furthermore, for the VAE-AE experiments the dimensionality of the latent vector is 10, for the VAE-MNIST it is 10, and for VAE-A3 it is 20. In all cases we also add the class dimensionality to the latent encoder vector to implement the conditional VAE variant.

\subsection{Association through Generalization Sleep Apnea: Model}
In this experiment we use the following  architecture:
\begin{itemize}
    \item  We use a convolutional network and we normalize and concatenate the 4 features to create an input dimensionality of 240 (i.e., $4\times60$). We use 3 convolutional layers ($1\times 8 \times 5\times 1$, MaxPool1D, $8 \times 16 \times 5\times 1$, MaxPool1D, $16 \times 32 \times 3\times 1$, MaxPool1D), and 3 fully connected layers (30$\cdot$32,480, $Out$). $Out$ depends on the model. $Out=8$ for the adversary. $Out=4$ for $h_{TU}$. $Out=2$ for the apnea detection classifier.  We use in all cases relu activations, and dropout in the 2 fully connected layers.
\end{itemize}

\subsection{Association through Generalization Sleep Stage: Models}

In this experiment we use the following  architectures:

\begin{itemize}
\item For models $h_T, h_S,D_{WGAN}, D_{VAE}$: We use a small MLP (Input,  256, 256, 128, 64, Output) with leaky relu activations and dropout of 0.75 in all layers. The output depends on the task and comprises 1 neuron for $D_{WGAN}$, dependent of experiment for $D_{VAE}$, 5 for $h_T,h_S$. Input is in all cases 58, i.e., equal to the input space's dimensionality $\mathbb X$.
\item For models $G_{WGAN}, G_{VAE}$: we use a similar inverted architecture as the above, i.e., (Input, 64, 128, 256, 256, Output=58). Again we use leaky relu and dropout of 0.75 in all layers. Furthermore,  Input is 10+5 for $D_{VAE}$, and 100+5 for $D_{WGAN}$).
\item For model $h_{TU}$, we use similar architecture as above, but with wider layers,i.e., (Input=58,512,512,256,128,Output=10). The Output 10 corresponds to the recordings of the open study.
\item For model $G_{AM}$ we use a smaller architecture than the other Generative models, since it can achieve better training times for AM and AM-I generation than the original generative architecture. We use the same dropout and activation conventions as before, with a fc architecture of (Input=100,128,64,64,Output=58).
\item Finally for adversary A-1 we use a wider but similar architecture as before, namely (Input=58, 1024, 1024, 512, 256, Output=21) since we observed that the wider architecture yielded better results for the  adversarial performance. A-2 is identical to A-1 but with 2 additional convolutional and Max-Pooling  layers at the beginning, namely (Input=58,1$\times$3$\times$1$\times$128, MP, 128$\times$3$\times$1$\times$256, MP, A-1).

\end{itemize}

\section{Adding Randomness}

In order to create a multi-faceted stimuli dataset, we would ideally need the AM distribution ($p_{AM}$) to have a uniform support in the regions which we have defined as important, based on our threshold of acceptance.

To achieve this we want (1) to capture a wide variety of initial positions from the feature space, and then perform AM   and (2) add variability to the AM process itself. To achieve these points we perform the following during the execution of our algorithm:
\begin{itemize}
    \item \textbf{Reinitializations:} We use our AM Generator to perform AM towards one class, we save the stimuli after it successfully surpassed the threshold of acceptance, and then we choose randomly another class to perform AM. Sporadically, we reinitialize the AM Generator, as the process can lead to numerical instability without reinitializations.
    
    \item \textbf{Random Thresholds:} In order to capture homogeneously the regions that are above a certain acceptance threshold ($T_{min}$) for the class probabilities of $h_T$, we use a randomly changing threshold with $T\in [T_{min},1]$.
    \item Use of\textbf{ larger initial weights} in the AM Generator: As we use reinitializations, we want to capture  a  large region of the feature space from the initial positions. For this reason when we use the AM Generator, we use larger than usual standard deviations in the weight initializations. However,  this can lead to numerical instability during the AM Maximization, so  consideration should be applied before using this strategy.
    
    \item \textbf{Use of edits instead of generation:} Instead of using a deep generator network $G_{AM}$, we randomly draw a sample $x_r$ from a random uniform distribution covering $\mathbb X$ (i.e., with support of $\mathbb X$). We then utilize a deep network $E_{AM}(,\theta_E, z)$ to perform edits on $x_r$ to modify it so that it can maximize an output neuron of $h_T$ based on $L_Y$. Since we assume that $\mathbb X$ is compact, we re-scale the total output to fit in $\mathbb X$ base on the min-max feature rescaling  (which we will call $resc$). As such, in summary we change the stimuli generation from $h_T(\theta^*_T,G_{AM}(\theta_G,z))$ to  $h_T(\theta^*_T,resc(x_r+E_{AM}(\theta_E,z)))$.  Empirically this technique yields good results regarding the abovementioned point (1). 
    
    \end{itemize}
    
    Additionally, another important detail to create a multi-faceted dataset relates to the output of $G_{AM}$. The output activation function is directly responsible for any hard constraints imposed in the stimuli. Though in some cases, like, e.g., the MNIST data, the form of the output can be clearly reproduced by the output activation function (e.g., rescaled pixel values 0-1 - sigmoid outputs), this is not always the case. All medical datasets for example are not as strictly ``bounded'' in their possible values. In most cases, the data values can be either positive or negative and  a pre-specified maximum or minimum value does not exist. As such, finding a proper function for the output of $G_{AM}$ is inherently harder. In most of the medical cases, we use a rescaled $tanh$, mutiplied by a constant occuring of the maximum and minimum values in the training data.

 \section{Closest Neighbors per Dataset}
   
\begin{figure}[h!]
\centering
  \includegraphics[scale=0.175]{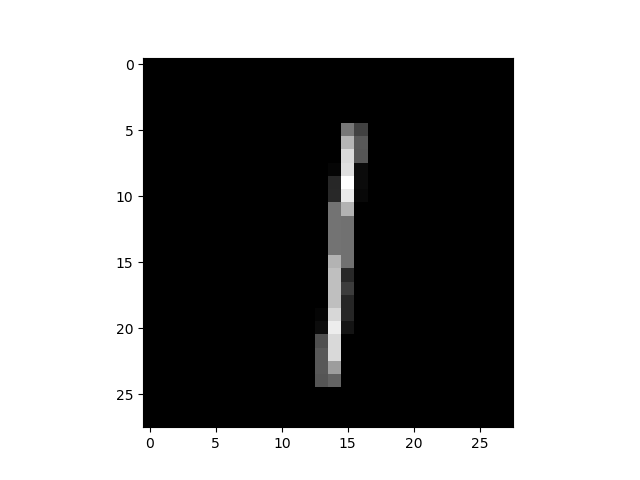}
  \includegraphics[scale=0.175]{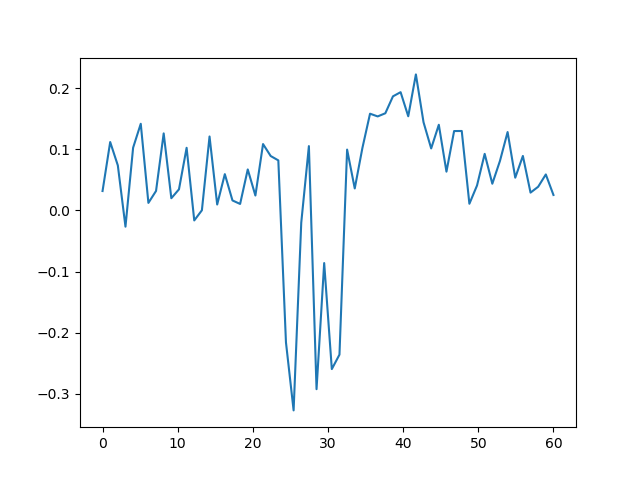}
  \includegraphics[scale=0.175]{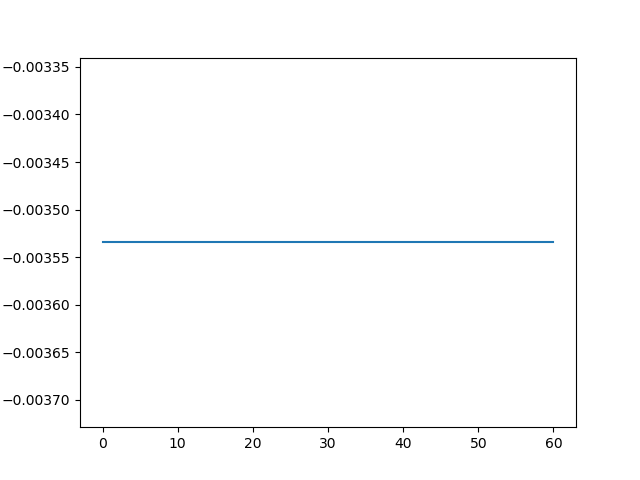}
 
  \includegraphics[scale=0.175]{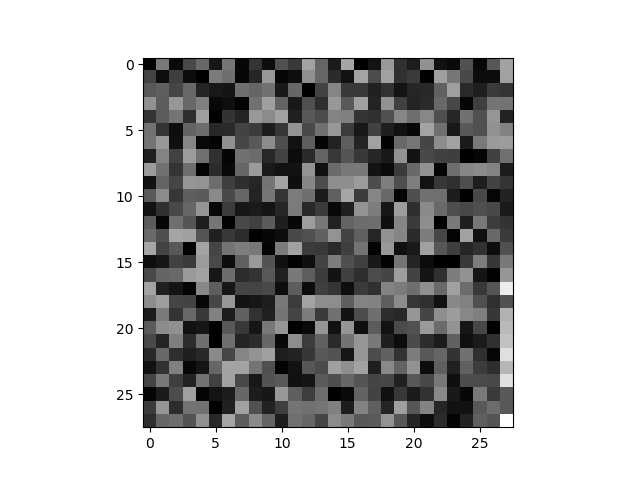}
  \includegraphics[scale=0.175]{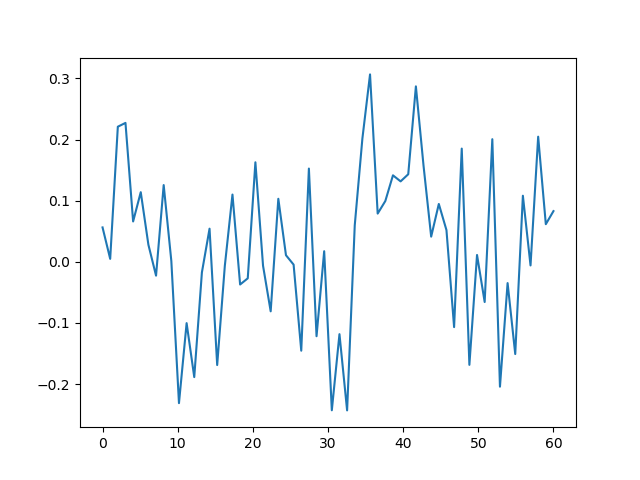}
  \includegraphics[scale=0.175]{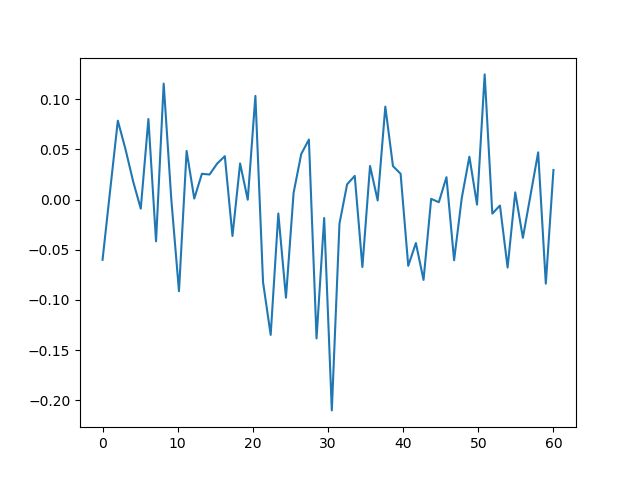}
\caption{ Closest neighbors between the Real(first row) and AM stimuli (second row)  datasets for MNIST (first column)  and Apnea-ECG(second column) A3 (third column)(mV). Since A3 is a real-world dataset, we have many ``inactive'' datapoints  for cases with sensor misplacement, bad sensor quality etc. The closest neighbor from the A3 study corresponds to such a datapoint.  }
\label{fig:Neighbors}
\end{figure}

 In this Appendix, we present the closest neighbors (l2-norm) between the real and AM datasets (Figure \ref{fig:Neighbors}).

\section{Theoretical Intuition: Proofs and Additional Notes}
In this Appendix we include the proofs for the lemmata of Section 2.5 and two small notes for the case where $R^r$ is of probability measure 0.

\subsection{Proofs}
 We include the proof for Lemma 2. Lemma 1  can be equivalently derived. For notational convenience we adopt: $\mathbb E[S^r_A(U,x)|x]= E_A(x)$.

 b): $\forall x \in R^r$ we identify two cases for a given adversary $A$:
 \begin{itemize}
     \item $A(x)=1$. Then $E_A(x)=\frac{p(u_t|x)}{p(u_t)}=\frac{p(u_t)}{p(u_t)}=1$
     
     \item $A(x)=0$. Then $E_A(x)=\frac{1-p(u_t|x)}{1-p(u_t)}=\frac{1-p(u_t)}{1-p(u_t)}=1$
 \end{itemize}

 a) For any ${A^r}^*\in \mathbf{A^r}^*$ and any $A'$, we identify the following cases:  
 
 \begin{itemize}
     \item $x\in R^r:$ Then from b), $ E_{{A^r}^*}(x)=E_{A'}(x)=1$
     \item $x\in R^r_+:$ Then if $A'(x)=1, E_{{A^r}^*}(x)=E_{A'}(x)= \frac{p(u_t|x)}{p(u_t)} $. Else, if $A'(x)=0$, we have that $E_{A'}(x)=\frac{1-p(u_t|x)}{1-p(u_t)}<1$, and $E_{{A^r}^*}(x)=\frac{p(u_t|x)}{p(u_t)}>1$. As such, $E_{A'}(x)<E_{{A^r}^*}(x)$
     
     \item $x \in R^r_-$: Equivalent with the above argument.
 \end{itemize}
   
 c):  Given ${A^r}^*\in \mathbf{A^r}^*$ we identify the following cases for any $x\not \in R^r$:
\begin{itemize}
     \item $x \in R^r_+$: Then ${A^r}^*(x)=1$ and  $\mathbb E_{{A^r}^*}(x)=\frac{p(u_t|x)}{p(u_t)}>1$
     
     \item  $x\in R^r_-$: Then ${A^r}^*(x)=0$ and  $E_{{A^r}^*}(x)=\frac{1-p(u_t|x)}{1-p(u_t)}>1$
 \end{itemize}

\subsection{Additional Notes}
 A question that arises from the previous analysis is what happens if region $R^r$ has measure 0 (e.g., $R^r$ is  a curve and $\mathbb X$ a surface).
 In this case it could  potentially be difficult to approximate $R^r$ with a gradient based method. 
 
\begin{table}[h!]
\small
\centering
\begin{tabular}{ | p{1.5cm}| p{1.5cm}| p{1.5cm}| p{1.5cm}| p{1.5cm}|}
\hline
\multicolumn{5}{|c|}{Hyper-Parameter values for AM, AM-I} \\
\hline
 &$T_Y$&$T_U$&$m$&$m'$\\
\hline
G:MNIST&0.9&-&20(50)&1(3,5)\\
G:AE&0.95&-&8&1\\
G:A3&0.7(0.6)&-&8&1\\
A:AE&0.95&0.5&8&1\\
A:Sl.St&0.5&0.4&64&1\\
\hline
\end{tabular}
\caption{ G: Generalization experiment (Section 4). A: Association through generalization (Section 5.1). AE is the Apnea-ECG, and Sl.St the Sleep Stage EDF data.}
 \label{table:ArchsT2}
\end{table}

\begin{table}[h!]
\small
\centering
\begin{tabular}{ | p{1cm}| p{2.5cm}| p{2.5cm}| p{1.5cm}| p{1.5cm}|}
\hline
\multicolumn{5}{|c|}{Median Maximum $l2$-distance for data within the class} \\
\hline
 Class&AM less restarts&AM static T&AM-base&Real\\
\hline
0&13.747&16.375&17.973&12.797\\
1&11.975&17.864&19.468&9.568\\
2&14.624&16.601&16.322&12.336\\
3&20.277&15.810&18.520&12.210\\
4&14.175&16.354&17.613&11.711\\
5&15.234&15.081&17.755&12.644\\
6&13.065&15.802&17.848&12.111\\
7&16.561&16.776&16.970&11.355\\
8&18.897&18.185&19.615&11.895\\
9&13.299&17.174&18.233&11.257\\
\hline
Mean&15.185&16.602&18.032&11.788\\
\hline
\end{tabular}
\caption{Median spread of the stimuli per-class}
 \label{table:diversity}
\end{table}

 However, we hypothesize that if we assume continuity of $p(u_t|x)$, it  is possible to show that we can  guarantee   score expectation that is close to 1  for all points in an extended  region  encompassing $R^r$, which can be defined as the union of  small neighborhoods near each point of $R^r$. This extended region has non-zero probability measure. For brevity and context consistency,  we leave the details of this analysis for future work.

 \section{Values of  AM, AM-I  Hyper-Parameters}
 In this Appendix we show the values used for the AM, AM-I Generation across the different experiments. In parentheses we show parameter values with which we obtain similar results as the main parameter values for the same amount of AMs (i.e.,the same amount of  inner loops in the algorithms). Threshold values show minimum threshold (i.e., we randomly choose a value between $[T_{U/Y},1)$).

 \section{Measuring Stimuli ``Spread'' Per-Class}
 
 In this Appendix we show the median maximum distance of the stimuli and the real data for each Class. For the stimuli, we additionally perform an ablation study where we do not perform several randomness operations. Specifically, we implement one version of  AM-generation where we use a fixed threshold of $T_Y=0.98$ instead of  drawing a random value from the $[0.96,0.99]$ range, as we did in our original design. Furthermore, we implement one version in which we increase the size of the candidate match $m'$ from 1 to 10 (i.e., more stimuli will be produced for the same gradient ascent of $G_{AM}$), and thus we perform less restarts to generate a pre-specified number of stimuli.

 We present our results in Table    \ref{table:diversity}. We observe that if we do not perform any ablation, we indeed get the largest values of median maximum distance for almost all Classes. Performing less restarts of $G_{AM}$ seems to have more of a negative effect on the data spread than using a static threshold. For all stimuli cases, the distance seems to be larger for the real data. As we do not aim to synthesize data that look visually realistic this is not problematic. On the contrary, as mentioned in Section \ref{sec:prelim}, we hypothesize that a larger ``spread'' within the input space could be indicative of higher variance in the stimuli data, which could prove to be useful as a way to compensate for the lack of realism within the dataset.  

 \section{Generalization and Customization: Accuracy, Specificity, and Sensitivity}

In this Appendix we present the Accuracy, Specificity and Sensitivity results from our experiments on the A3 and the Apnea-ECG datasets (Tables \ref{tableApnea:Accuracy}, \ref{tableApnea:Sensitivity}, \ref{tableApnea:Specificity}, \ref{tableApnea:Precision}).

\begin{table}[h]
\centering
\footnotesize
\begin{tabular}{ | p{1.5cm}| p{1.5cm}|p{1.5cm}|p{1.5cm}|p{1.5cm}|}
\hline
 &Baseline & VAE & WGAN &AM  \\
\hline
AE ID:&94.36$\pm$0.22&74.04$\pm$1.07&90.26$\pm$0.37&92.96$\pm$0.21\\
AE S: &94.24$\pm$0.24&74.31$\pm$1.64&89.72$\pm$0.38&91.61$\pm$0.29\\
AE VS: &93.78$\pm$0.28&72.85$\pm$1.25&89.03$\pm$0.87&91.98$\pm$0.31\\
AE L: &93.62$\pm$0.26&74.63$\pm$1.23&89.86$\pm$0.32&92.07$\pm$0.25\\
AE LL:&94.11$\pm$0.23&76.19$\pm$1.43&88.20$\pm$0.31&91.38$\pm$0.26\\
\hline
\hline
A3 ID:&86.40$\pm$0.12&\qquad -&82.58$\pm$0.19&85.21$\pm$0.10\\
A3 S: &86.67$\pm$0.02&\qquad -&82.69$\pm$0.19&84.96$\pm$0.08\\
A3 VS:&86.30$\pm$0.03&\qquad -&81.95$\pm$0.40&84.54$\pm$0.11\\
A3 L: &86.64$\pm$0.07&\qquad -&80.93$\pm$1.00& 85.28$\pm$0.08\\
A3 LL:&86.84$\pm$0.02&\qquad -&82.47$\pm$0.18&84.68$\pm$0.13\\
\hline
\end{tabular}
\caption{Accuracy results for the Apnea generalization and customization experiments for Apnea-ECG and A3 datasets} 
 \label{tableApnea:Accuracy}
\end{table}

\begin{table}[h]
\centering
\footnotesize
\begin{tabular}{ | p{1.5cm}| p{1.5cm}|p{1.5cm}|p{1.5cm}|p{1.5cm}|}
\hline
 &Baseline & VAE&WGAN &AM  \\
\hline
AE ID:&94.62$\pm$0.36&88.77$\pm$1.23&88.91$\pm$0.66&93.52$\pm$0.43\\
AE S: &94.16$\pm$0.33&87.07$\pm$0.17&90.39$\pm$0.79&92.25$\pm$0.55\\
AE VS: &94.05$\pm$0.43&85.70$\pm$1.82&91.37$\pm$0.90&90.84$\pm$0.52\\
AE L:&93.05$\pm$0.46&92.20$\pm$1.08&91.59$\pm$0.59&93.77$\pm$0.38\\
AE LL:&93.11$\pm$0.36&90.83$\pm$1.54&89.92$\pm$0.81&89.29$\pm$0.62\\
\hline
\hline
A3 ID:&65.10$\pm$0.27&\qquad -&69.61$\pm$1.06&64.64$\pm$1.03\\
A3 S: &64.01$\pm$0.16&\qquad -&69.23$\pm$0.81&64.35$\pm$1.21\\
A3 VS: &60.86$\pm$0.19&\qquad -&68.68$\pm$1.01&62.41$\pm$1.28\\
A3 L: &65.65$\pm$0.19&\qquad -&69.82$\pm$1.95&63.88$\pm$0.91\\
A3 LL:&64.94$\pm$0.35&\qquad -&67.03$\pm$0.89&54.19$\pm$1.42\\
\hline
\end{tabular}
\caption{Sensitivity results } 
 \label{tableApnea:Sensitivity}
\end{table}

\begin{table}[h]
\centering
\footnotesize
\begin{tabular}{ | p{1.5cm}| p{1.5cm}|p{1.5cm}|p{1.5cm}|p{1.5cm}|}
\hline
 &Baseline & VAE& WGAN &AM  \\
\hline
AE ID:&94.18$\pm$0.30&63.83$\pm$2.00&91.21$\pm$0.42&92.57$\pm$0.33\\
AE S: &94.30$\pm$0.27&65.33$\pm$3.21&89.27$\pm$0.53&91.18$\pm$0.38\\
AE VS: &93.58$\pm$0.35&63.96$\pm$2.15&87.43$\pm$0.84&92.73$\pm$0.48\\
AE L:&94.00$\pm$0.43&62.122.42&88.66$\pm$0.41&90.96$\pm$0.41\\
AE LL:&94.77$\pm$0.31&66.27$\pm$2.33&87.07$\pm$0.52&92.84$\pm$0.34\\
\hline
\hline
A3 ID:&93.43$\pm$0.13&\qquad-&86.85$\pm$0.45&92.00$\pm$0.29\\
A3 S: &94.14$\pm$0.05&\qquad -&87.13$\pm$0.50&91.75$\pm$0.44\\
A3 VS: &94.70$\pm$0.03&\qquad -&86.32$\pm$0.81&91.85$\pm$0.37\\
A3 L: &93.56$\pm$0.07&\qquad -&84.60$\pm$1.81&92.34$\pm$0.33\\
A3 LL:&94.07$\pm$0.10&\qquad -&87.56$\pm$45&94.74$\pm$0.35\\
\hline
\end{tabular}
\caption{Specificity results } 
 \label{tableApnea:Precision}
\end{table}

\begin{table}[h!]
\centering
\footnotesize
\begin{tabular}{ | p{1.5cm}| p{1.5cm}|p{1.5cm}|p{1.5cm}|p{1.5cm}|}
\hline
 &Baseline & VAE&WGAN &AM  \\
\hline
AE ID:&91.56$\pm$0.40&63.29$\pm$1.25&87.45$\pm$0.55&89.60$\pm$0.38\\
AE S: &91.72$\pm$0.41&64.47$\pm$1.44&85.23$\pm$0.67&87.73$\pm$0.50\\
AE VS: &91.17$\pm$0.45&62.42$\pm$1.46&83.69$\pm$0.87&89.46$\pm$0.65\\
AE L:&91.61$\pm$0.49&64.08$\pm$1.41&84.87$\pm$0.52&87.46$\pm$0.59\\
AE LL:&92.43$\pm$0.37&65.15$\pm$1.67&82.20$\pm$0.62&89.47$\pm$0.49\\
\hline
\hline
A3 ID:&76.61$\pm$0.38&\qquad-&63.69$\pm$0.55&72.82$\pm$0.47\\
A3 S: &78.31$\pm$0.12&\qquad-&64.09$\pm$0.63&72.23$\pm$0.75\\
A3 VS: &79.12$\pm$0.08&\qquad-&62.67$\pm$1.12&71.80$\pm$0.60\\
A3 L: &77.11$\pm$0.20&\qquad-&60.97$\pm$1.04&73.48$\pm$0.63\\
A3 LL:&78.35$\pm$0.212&\qquad-&64.11$\pm$0.60&77.52$\pm$0.75\\
\hline
\end{tabular}
\caption{Precision results} 
 \label{tableApnea:Specificity}
\end{table}

\section{Steps 1,3, and 4 of the Proposed Approach}

In this Appendix, we discuss the additional steps needed for the AM-generation.

\begin{itemize}
    \item \textbf{Step 1: } We want to train $h_T$ with a labelled dataset.  Supervised learninig is commonly used for such tasks. As on Step 2 we will subject $h_T$ to gradient-based Activation Maximization, we want $h_T$ to be a model with which such procedure can occur. Neural Nets are examples of such models. Apart from this, there are no other hyperparameters or design constraints.  Any training Algorithm, Optimizer, etc can be used.
    
    \item \textbf{Step 3: } Similar to Step 1, but in this case we have no constrain about the Model ($h_S$ in this case). We train $h_S$ on the stimuli and the  labels occuring from the generation proccess of Step 2 in a supervised manner. As such any Algorithm, design choice and hyper-parameter tuning  can be used. However generally, for DNNs and CNNs as $h_T$ we observed that $h_S$ yielded better generalization performance when it was similar to $h_T$ (Architecturally and Algorithmically),

    \item \textbf{Step 4:}  Use $h_S$ to perform inference on data from the same domain as the original data. 
\end{itemize}
\vskip 0.2in
\bibliography{mybibfile}
\bibliographystyle{theapa}

\end{document}